\documentclass[journal]{IEEEtran}
\usepackage{cite,footnote,xspace,syntonly}
\usepackage{amsmath,mathtools}

\usepackage{fixltx2e}

\usepackage{graphicx}

\usepackage{subcaption}              

\usepackage[utf8]{inputenc}

\usepackage{amsfonts}
\usepackage{amsmath}
\usepackage{mathtools}

\usepackage{amssymb}

\usepackage{bm}

\usepackage{color,verbatim}
\usepackage{multirow}
\usepackage{accents}

\usepackage{theoremref}

\usepackage{url}

\newcounter{rulecounter}
\newcommand{\resetrule}{ \setcounter{rulecounter}{0}}
\resetrule

\newsavebox{\selvestebox}
\newenvironment{colbox}[1]
  {\newcommand\colboxcolor{#1}%
   \begin{lrbox}{\selvestebox}%
   \begin{minipage}{\dimexpr\columnwidth-2\fboxsep\relax}}
  {\end{minipage}\end{lrbox}%
   \begin{center}
   \colorbox{\colboxcolor}{\usebox{\selvestebox}}
   \end{center}}

\definecolor{orange}{rgb}{1,0.8,0}
\definecolor{gray}{rgb}{.9,0.9,0.9}
\definecolor{darkgray}{rgb}{.3,0.3,0.3}
\definecolor{darkblue}{rgb}{.1,0.0,0.3}
\definecolor{lightblue}{rgb}{0.7,0.7,1}
\definecolor{lightred}{rgb}{1,0.7,.7}
\definecolor{purple}{RGB}{204,153,255}
\definecolor{lightgray}{rgb}{.95,0.95,0.95}
\definecolor{lightgreen}{rgb}{0.3,0.5,0.3}
\definecolor{darkgreen}{rgb}{0.05,0.3,0.05}

\newcommand{\ra}{$\rightarrow$~}

\newcommand{\brackets}[1]{\left\{#1\right\}}
\newcommand{\bbm}[1]{{\bar{\bm #1}}}

\newcommand{\hbm}[1]{{\hat{\bm #1}}}
\newcommand{\inv}{^{-1}}
\newcommand{\pinv}{^{\dagger}}

\newcommand{\rfield}{\mathbb{R}}
\newcommand{\vect}{\mathop{\rm vec}}
\newcommand{\vectinv}{\mathop{\rm vec}^{-1}}

\newcommand{\diag}[1]{\mathop{\rm diag}\brackets{#1}}

\newcommand{\bdiag}[1]{\mathop{\rm bdiag}\brackets{#1}}
\newcommand{\diagnb}{\mathop{\rm diag}}

\newcommand{\tr}[1]{\mathop{\rm Tr}\left(#1\right)}

\newcommand{\transpose}{^T}
 \newcommand{\define}{\triangleq}

\newcommand{\expectednb}{\mathop{\textrm{E}}\nolimits}

\DeclareMathOperator*{\argmin}{arg\,min}

\newtheorem{myproposition}{Proposition}
\newtheorem{myremark}{Remark}
\newtheorem{myproblemstatement}{Problem Statement}
\newtheorem{mylemma}{Lemma}
\newtheorem{mytheorem}{Theorem}
\newtheorem{mydefinition}{Definition}
\newtheorem{mycorollary}{Corollary}

\newcounter{exampleind}

\newcounter{remarkind}

\renewcommand{\vectinv}{\mathop{\rm unvec}}

\newcommand{\bdiagnb}{\mathop{\rm bdiag}}
\newcommand{\extendedfactor}{\hc{\mathcal{\bbm T}}}
\newcommand{\extendedfactornoobs}{\hc{\mathcal{\bbm T}'}}
\newcommand{\selectormatrix}{\hc{\mathcal{\bbm J}}}
\newcommand{\extendedobs}{\hc{\mathcal{\bbm A}}}
\newcommand{\extendedtrs}{\hc{\mathcal{\bbm A}'}}
\newcommand{\extendedsigma}{\hc{\bbm{\Sigma}}}
\newcommand{\extendednoobssigma}{\hc{\bbm{\Sigma}'}}
\newcommand{\extendedgrowingobsvec}{\hc{\bbm{\psi}}}
\newcommand{\extendedgrowingnoobsvec}{\hc{\bbm{\psi}'}}
\newcommand{\verticalincrement}{\hc{\bm{V}}}
\newcommand{\horizontalincrement}{\hc{\bm{W}}}
\newcommand{\extendedgrowingsignalestvec}{\hc{\hat{\bar{\bm \signalfun}}}} 
\newcommand{\extendedgrowingnoisevec}{\hc{\bar{\bm \epsilon}}} 
\newcommand{\extendedgrowingsignalvec}{\hc{\bbm \signalfun}}
\newcommand{\canonicalvec}[2]{{\bm i}_{#1,#2}}
\newcommand{\tridiag}{\mathop{\rm btridiag}}

\renewcommand{\define}{:=}

\renewcommand{\expectednb}{\hc{\mathbb{E}}}

\newcommand{\pdset}{\hc{\mathbb{S}}_+}

\newcommand{\timenot}[1]{{\hc{[}{#1}\hc{]}}}  %
\newcommand{\timegiventimenot}[2]{{\hc{[}{#1}|{#2}\hc{]}}}  %
\newcommand{\giventimenot}[1]{{\hc{\big|}{#1}\hc{]}}}  %
\newcommand{\timegiventimevertexnot}[3]{_{#3}{\hc{[}{#1}|{#2}\hc{]}}}  %
\newcommand{\timeind}{{\hc{t}}} 
\newcommand{\timeindaux}{{\hc{\tau}}} 
\newcommand{\timeindp}{{\hc{\timeind}'}} 
\newcommand{\timenum}{{\hc{T}}} 
\newcommand{\timeset}{\hc{\mathcal{T}}} 
\newcommand{\vertexnot}[1]{_{#1}}  %
\newcommand{\timevertexnot}[2]{_{#2}{\hc{[}{#1}\hc{]}}}  %
\newcommand{\vertexvertexnot}[2]{_{#1,#2}}  %
\newcommand{\timevertexvertexnot}[3]{_{#2,#3}{\hc{[}{#1}\hc{]}}}  %
\newcommand{\sampletimenot}[2]{_{#1}{\hc{[}{#2}\hc{]}}}  %

\newcommand{\graph}{\hc{\mathcal{G}}}
\newcommand{\extendedgraph}{\hc{\bar{\mathcal{G}}}}
\newcommand{\vertexset}{\hc{\mathcal{V}}}
\newcommand{\extendedvertexset}{\hc{\bar{\vertexset}}}
\newcommand{\edgeset}{\hc{\mathcal{E}}}
\newcommand{\vertexind}{{\hc{{n}}}}

\newcommand{\vertexindp}{{\hc{{\vertexind}'}}} 
\newcommand{\vertexnum}{{\hc{{N}}}}

\newcommand{\extendedadjacencymat}{\hc{\bbm A}} 
\newcommand{\timeconnectionmat}{\hc{\bm B}\alongtime} 
\newcommand{\timeconnection}{\hc{b}\alongtime} 
\newcommand{\timeconnectionmatcolsumvec}{\hc{\bm b}\alongtime} 

\newcommand{\extendedlaplacianmat}{\hc{\bbm L}}

\newcommand{\laplacianeval}{\hc{\lambda}} 
 
\newcommand{\alongspace}{_{\vertexset}} 
\newcommand{\alongtime}{_{\timeset}} 
\newcommand{\spaceadjacencymat}{\hc{\bm A}\alongspace} 
\newcommand{\spaceadjacencymatentry}{\hc{A}^{\vertexset}} 
\newcommand{\spacelaplacianmat}{\hc{\bm L}\alongspace} 
\newcommand{\timeadjacencymat}{\hc{\bm A}\alongtime} 
\newcommand{\timelaplacianmat}{\hc{\bm L}\alongtime} 
\newcommand{\timelaplacianevecmat}{\hc{\bm U}\alongtime} 
\newcommand{\timelaplacianeval}{\hc{ \lambda}\alongtime} 
\newcommand{\timelaplacianevalvec}{\hc{\bm \lambda}\alongtime} 
\newcommand{\spacelaplacianevecmat}{\hc{\bm U}\alongspace} 
\newcommand{\spacelaplacianeval}{\hc{\lambda}\alongspace}
\newcommand{\spacelaplacianevalvertexnot}[1]{\hc{\lambda}^{\vertexset}_{#1}} 
\newcommand{\spacelaplacianevec}{\hc{\bm u}^{\vertexset}} 
\newcommand{\spacelaplacianevalvec}{\hc{\bm \lambda}\alongspace}

\newcommand{\signalfun}{{\hc{f}}}
\newcommand{\extendedsignalfun}{{\hc{\bar f}}}

\newcommand{\signalvec}{\hc{\bm \signalfun}} 
\newcommand{\extendedsignalvec}{\hc{\bbm \signalfun}} 
\newcommand{\extendedsignalmat}{\hc{\bbm F}} 

\newcommand{\fourierextendedsignalmat}{\hc{\fourier{\bbm F}}} 
\newcommand{\fourierextendedsignalvec}{\hc{\fourier{\bbm \signalfun}}} 

\newcommand{\fouriersignalfun}{\hc{\fourier f}} 
\newcommand{\fouriersignalvec}{\hc{\fourier{\signalvec}}}
\newcommand{\signalestfun}{\hc{\hat \signalfun}} 
\newcommand{\signalestvec}{\hc{\hbm \signalfun}}

\newcommand{\extendedsignalestvec}{\hc{\hat{\bbm \signalfun}}}

\newcommand{\sampleset}{\hc{\mathcal{S}}} 
\newcommand{\samplemat}{\hc{\bm S}} 
\newcommand{\extendedsamplemat}{\hc{\bar{\samplemat}}} 
\newcommand{\sampleind}{{\hc{s}}} 
\newcommand{\samplenum}{{\hc{S}}} 
\newcommand{\samplenumdiagmat}{{\hc{\bm D_{\samplenum}}}} 
\newcommand{\extendedsamplenum}{{\hc{\bar S}}} 
\newcommand{\observationfun}{{\hc{y}}}

\newcommand{\observationvec}{\hc{\bm y}} 
 
\newcommand{\extendedobservationvec}{\hc{\bbm y}}

\newcommand{\observationnoisefun}{{\hc{e}}} 
\newcommand{\observationnoisevec}{{\bm\observationnoisefun}}

\newcommand{\observationnoisevar}{\hc{ {\sigma^2_\observationnoisefun}}}  %
\newcommand{\rkhsfunsymbol}{f}
\newcommand{\rkhsvec}{\hc{{\bm \rkhsfunsymbol}}}
\newcommand{\extendedrkhsvec}{\hc{{\bbm \rkhsfunsymbol}}}

\newcommand{\rkhsestvec}{\hc{\hbm \rkhsfunsymbol}}

\newcommand{\fullkernelmat}{\hc{\bm K}} 
\newcommand{\spacefullkernelmat}{\fullkernelmat\alongspace} 
\newcommand{\timefullkernelmat}{\fullkernelmat\alongtime} 
 
\newcommand{\kernelinvondiagonalmat}{\hc{\bm D}} 
\newcommand{\kernelinvoffdiagonalmat}{\hc{\bm C}} 

\newcommand{\extendedfullkernelmat}{\hc{\bbm K}}

\newcommand{\frequencyweightfun}{\hc{r}}
\newcommand{\spacefrequencyweightfun}{\hc{r}\alongspace}
\newcommand{\timefrequencyweightfun}{\hc{r}\alongtime}
\newcommand{\frequencyweightvec}{\hc{\bm r}}
\newcommand{\frequencyweightmat}{\hc{\bm R}}
 
\newcommand{\blnum}{{\hc{B}}} 
\newcommand{\transitionmat}{{\hc{\bm P}}}
\newcommand{\errormat}{{\hc{\bm M}}}
\newcommand{\plantnoise}{w} 
 
\newcommand{\plantnoisevec}{\hc{\bm \plantnoise}} 

\newcommand{\plantnoisekernelmatentry}{{\hc{\Sigma}}} 
\newcommand{\plantnoisekernelmat}{{\hc{\bm \plantnoisekernelmatentry}}} 
\newcommand{\kalmangainmat}{{\hc{\bm G}}}

\newcommand{\regpar}{\hc{\mu}}
\newcommand{\regfun}{\hc{\rho}}
\newcommand{\graphvariation}{\regfun_\text{LR}}
\newcommand{\laplaciankernelregfun}{\regfun_\text{LK}}

\newcommand{\fourier}[1]{\check{#1}}
\newcommand{\frequencytimeind}{\fourier{\timeind}}
\newcommand{\frequencyvertexind}{{\fourier{\vertexind}}}

\usepackage{epstopdf}
\usepackage{algorithmic}
\usepackage[ruled,vlined,resetcount]{algorithm2e}

\usepackage{amsthm}

\theoremstyle{plain}
\newtheorem{mylemmahere}{Lemma}
\newtheorem{mytheoremhere}{Theorem}
\newtheorem{mydefinitionhere}{Definition}
\theoremstyle{definition}
\newtheorem{myremarkhere}{Remark}

\newcommand{\change}[1]{\textcolor{black}{#1}}

 \newcommand{\cmt}[1]{} 
 \newcommand{\hc}[1]{{#1}} 
\newenvironment{myitemize}{}{}
\newcommand{\myitem}{}

\renewcommand{\transpose}{^{\hc{\top}}}

\newcommand{\identitymat}{\hc{\bm I}}

\begin{document}
\title{Kernel-based Reconstruction  of\\  Space-time Functions on Dynamic Graphs} %
\author{Daniel Romero, \emph{Member, IEEE}, Vassilis N. Ioannidis, Georgios
  B. Giannakis, \emph{Fellow, IEEE} \thanks{This work was supported by
    ARO grant W911NF-15-1-0492 and NSF grants 1343248, 1442686, and
    1514056.

 The authors are with the Dept. of ECE and the Digital Tech. Center,
 Univ. of Minnesota, USA. E-mail: \{dromero,ioann006,georgios\}@umn.edu}}
\maketitle

\begin{abstract}
Graph-based methods pervade the inference toolkits of numerous
disciplines including sociology, biology, neuroscience, physics,
chemistry, and engineering.  A challenging problem encountered in this
context pertains to determining the attributes of a set of vertices
given those of another subset at possibly different time
instants. Leveraging spatiotemporal dynamics can drastically reduce
the number of observed vertices, and hence the cost of sampling.
Alleviating the limited flexibility of existing approaches, the
present paper broadens the existing kernel-based graph function
reconstruction framework to accommodate time-evolving functions over
possibly time-evolving topologies. This approach inherits the
versatility and generality of kernel-based methods, for which no
knowledge on distributions or second-order statistics is required.
Systematic guidelines are provided to construct two families of
space-time kernels with complementary strengths. The first facilitates
judicious control of regularization on a space-time frequency plane,
whereas the second can afford time-varying topologies.  Batch and
online estimators are also put forth, and a novel kernel Kalman filter
is developed to obtain these estimates at affordable computational
cost. Numerical tests with real data sets corroborate the merits of
the proposed methods relative to competing alternatives.
\end{abstract}

\section{Introduction}
\label{sec:intro}

\cmt{motivation time-inv}
\begin{myitemize}

\myitem\cmt{graph data}A number of applications involving social,
biological, brain, sensor, transportation, or communication networks
call \myitem\cmt{infer attr.}for efficient methods to infer the
attributes of some vertices given the attributes of other
vertices~\cite{kolaczyck2009}.  \myitem\cmt{example: social nets}For
example, in a social network with vertices and edges respectively
representing persons and friendships, one may be interested in
determining an individual's consumption trends based on those of their
friends.  \myitem\cmt{partial observability}This task emerges when
sampling cost constraints, such as the impossibility to poll one
country's entire population about political orientation, limit
the number of vertices with known attributes.
\myitem\cmt{formalized}Existing approaches typically formulate this
problem as the reconstruction of a function or signal on a
graph~\cite{kondor2002diffusion,smola2003kernels,kolaczyck2009,shuman2013emerging,sandryhaila2013discrete,chapelle2006},
\myitem\cmt{parsimony}and rely on its smoothness with respect to the
graph, in the sense that neighboring vertices have similar function
values.  \myitem\cmt{example:age}This principle suggests, for
instance, estimating one person's age by looking at their friends'
age.

\end{myitemize}

\cmt{Time series reconstr.}
\begin{myitemize}
\myitem\cmt{time series motiv.}A more challenging problem involves
reconstructing time-evolving functions on graphs,
\cmt{example:brain}such as the ones describing the time-dependent
activity of regions in a brain network, given their values on a subset
of vertices and time instants. \myitem\cmt{Impact}Efficiently
exploiting spatiotemporal dynamics can markedly impact sampling costs
by reducing the number of vertices that need to be observed to attain
a target performance. \myitem\cmt{example:ECoG}Such a reduction is of
paramount interest in applications such as invasive
electrocorticography (ECoG), where observing a vertex requires the
implantation of an intracranial electrode~\cite{kramer2008seizure}.
\end{myitemize}

\cmt{Literature:time-invariant funs}An extensive body of literature
has dealt with reconstructing time-invariant graph functions.
\begin{myitemize}
\myitem\cmt{Machine learning}Machine learning works typically rely on
smoothness~\cite{chapelle2006,kondor2002diffusion,smola2003kernels,belkin2006manifold}
to reconstruct \cmt{mainly classification}either binary-valued (see
e.g.~\cite{chapelle2006}) or real-valued
functions~\cite{chapelle1999transductive,belkin2006manifold,cortes2007transductive,lafferty2007regression},
\myitem\cmt{SPoG}whereas the community of signal processing on graphs
(SPoG) focuses on {parametric} estimators \cmt{real-val}for
real-valued functions adhering to the
\cmt{bandlimitedness}\emph{bandlimited model}, by which those
functions are confined to the span of $\blnum$ eigenvectors of the
graph Laplacian or adjacency
matrices~\cite{tsitsvero2016uncertainty,anis2016proxies,narang2013localized,wang2015local,marques2015aggregations}. 
\myitem\cmt{Meng's paper}Most of these schemes can be subsumed under the
encompassing framework of time-invariant kernel-based
learning~\cite{romero2016multikernel}.
\end{myitemize}

\cmt{Literature:time-varying funs}\cmt{motiv}
\begin{myitemize}
  \myitem\cmt{time dependence explicitly modeled}Schemes tailored for
  time-evolving  functions on graphs 
  \begin{myitemize}
    \myitem\cmt{\cite{bach2004learning} (bach and jordan) and
      \cite{mei2016causal} (Mei and
      Moura)}include~\cite{bach2004learning} and \cite{mei2016causal},
    which 
    \begin{myitemize}
      \myitem\cmt{goal}predict the function values at time
      $\timeind$ given observations up to time $\timeind-1$.
      \myitem\cmt{limit.}However, these schemes assume that
      the function of interest adheres to a specific vector
      autoregression and all vertices are observed at previous
      time instances.  Moreover, \cite{bach2004learning} requires
      Gaussianity along with a rather \emph{ad hoc} form of stationarity.
    \end{myitemize}
  \end{myitemize}

  \myitem\cmt{time dependence unmodeled}Other works target
  time-invariant functions, but can afford tracking sufficiently slow
  variations. This is the case of the dictionary learning approach
  in~\cite{forero2014dictionary} and the distributed algorithms
  in~\cite{wang2015distributed} and
  \cite{lorenzo2016lms}. Unfortunately, the flexibility of these
  algorithms to capture spatial information is also limited
  since~\cite{forero2014dictionary} focuses on Laplacian
  regularization, whereas~\cite{wang2015distributed} and
  \cite{lorenzo2016lms} require the signal to be bandlimited.
  \myitem\cmt{algs. for apps}
  \begin{myitemize}    
    \myitem\cmt{\cite{rajawat2014cartography}
      (Rajawat-Emiliano-GG)\&\cite{kekatos2014electricity}
      (Kekatos-Zhang-GG)}Different approaches investigate special
    instances of the reconstruction problem with domain-specific
    requirements and
    assumptions~\cite{rajawat2014cartography,kekatos2014electricity}.
  \end{myitemize}
Finally, it is worth mentioning that no approach deals with
time-evolving topologies.


\end{myitemize}

\cmt{Contributions}The contribution of this paper is threefold. 
\begin{myitemize}
  \myitem\cmt{generalized framework}First, the existing kernel-based
  learning framework is naturally extended to subsume time-evolving
  functions over possibly dynamic graphs through the notion of
  \emph{graph extension}, by which the time dimension receives the
  same treatment as the spatial dimension.  The versatility of
  kernel-based methods to leverage spatial
  information~\cite{romero2016multikernel} is thereby inherited and
  broadened to account for temporal dynamics as well.  Incidentally,
  this vantage point also accommodates time-varying sampling sets and
  topologies.
  \myitem\cmt{kernels}Second, two families of \emph{space-time
  kernels} are introduced by generalizing  Laplacian
  kernels~\cite{smola2003kernels}. The first family enables kernel
  design in a bidimensional frequency domain, whereas the second caters
  for time-varying topologies.
  \myitem\cmt{KRR estimators}The third contribution comprises two
  \emph{function estimators} with complementary strengths  based on the
  popular kernel ridge regression (KRR) criterion; see
  e.g.~\cite{scholkopf2002,romero2016multikernel}. Whereas the first
  can handle more sophisticated forms of spatiotemporal
  regularization, the second can afford a more efficient implementation
  and online operation, meaning that estimates are refined as
  observations become available. The proposed kernel Kalman filter
  (KKF) finds exact online KRR estimates by implicitly operating in a
  (possibly) infinite-dimensional space.

\end{myitemize}

\cmt{Novelty}The major novelty of this paper is a
\begin{myitemize}
  \myitem\cmt{Deterministic methodology}purely deterministic
  methodology that obviates the need for assumptions on data
  distributions, stationarity, or knowledge of second-order
  statistics. The proposed schemes are therefore of special interest
  in absence of sufficient historical data, yet the latter can be
  incorporated if available through covariance
  kernels~\cite{romero2016multikernel}. Although more complicated
  dynamics can be accommodated, one may simply rely on the assumption
  that the target function is smooth over the graph and over time,
  which is reasonable whenever the graph is properly constructed and
  the sampling interval is attuned to the temporal dynamics of the
  function.  \myitem\cmt{kernel KF}The novel online estimator
  constitutes the first \change{fully deterministic} rigorous
  application of the Kalman filter (KF) to kernel-based learning.
  \begin{myitemize}
    \myitem\cmt{\cite{ralaivola2005timeseries}(Ralaivola)}Although~\cite{ralaivola2005timeseries}
    already proposed a kernel-based KF, this work heavily relies on
    heuristics and approximations to explicitly operate in feature
    space. Moreover, this algorithm involves solving the challenging
    preimage problem per time step, which increases inaccuracy and
    computational cost.
    \myitem{}\cmt{\cite{zhu2014generative}}\change{Another KF was
      developed in \cite{zhu2014generative} within
      the framework of kernel-based learning, but its formulation is
      probabilistic and requires historical data to estimate the data
      distribution.}
  \end{myitemize}

\end{myitemize}

\cmt{Paper structure overview}The rest of the paper is structured as
follows. Sec.~\ref{sec:problemdef} formulates the problem
and~Sec.~\ref{sec:background} reviews kernel-based learning for
time-invariant functions.  Sec.~\ref{sec:reconstruction} generalizes
this framework to reconstruct time-evolving functions and develops two
estimators together with the KKF. Space-time kernels are designed in
Sec.~\ref{sec:kernels}.  The numerical tests in
Sec.~\ref{sec:simulatedtests} confirm the benefits of the proposed
algorithms. Finally, Sec.~\ref{sec:conclusions}
summarizes closing remarks whereas the Appendix provides the proofs of the
main results.

\cmt{Notation}\noindent\emph{Notation:}
\begin{myitemize}
  \myitem  Scalars are denoted by
lowercase letters, vectors by bold lowercase, and matrices by bold
uppercase. 
  \myitem $(\bm A)_{m,n}$ is the  $(m,n)$-th
  entry of matrix $\bm A$.
  \myitem  Superscripts $~\transpose$  and   $~\pinv$  respectively
  denote transpose and pseudo-inverse.
  \myitem If $\bm A\define[\bm a_1,\ldots,\bm a_N]$, then $\vect\{\bm
  A\}
\define [\bm a_1\transpose,\ldots,\bm a_N\transpose]\transpose\define\bm a$
  and  $\vectinv\{\bm a\}\define\bm A$.
  \myitem With $\vertexnum\times\vertexnum$ matrices $\{\bm A_\timeind\}_{\timeind=1}^\timenum$
  and $\{\bm B_\timeind\}_{\timeind=2}^\timenum$ with $\bm A_\timeind
  = \bm A_\timeind\transpose~\forall\timeind$, 
$ \tridiag\{\bm A_1,\ldots,\bm A_\timenum;\bm B_2,\ldots,\bm B_\timenum\}$
represents the symmetric block tridiagonal matrix
  \begin{align*}
\left[ 
    \begin{array}{cccccc}
      \bm A_{1}  &
      \bm B\transpose _{2} &\bm0 &\ldots & \bm0 & \bm0 \\
      \bm B_{2} & \bm A_{2} &\bm B\transpose_{3} & \ldots &\bm0 & \bm0 \\
      \bm0 & \bm B_{3} & \bm A_{3} &\ldots & \bm0 & \bm 0\\
      \vdots & \vdots & \vdots & \ddots & \vdots & \vdots\\
      \bm0 & \bm0 & \bm0 &\ldots &\bm A_{\timenum-1}& \bm B\transpose_{\timenum}\\
      \bm0 & \bm0 & \bm0 &\ldots & \bm B_{\timenum} &\bm A_{\timenum}\\
    \end{array}
  \right].
\end{align*}
  \myitem Similarly, $\bdiag{\bm A_1,\ldots,\bm A_N}
  \define\tridiag\{\bm A_1,\ldots,$ $\bm A_N;\bm 0,\ldots,\bm 0\}$ is
  a block diagonal matrix.  \myitem Symbols $\odot$, $\otimes$, and
  $\oplus$ respectively denote element-wise (Hadamard) matrix product,
  Kronecker product, and Kronecker sum, the latter being defined for
  $\bm A\in \rfield^{M \times M}$ and $\bm B\in \rfield^{N \times N}$
  as $\bm A \oplus \bm B \define \bm A \otimes \bm I_N + \bm I_M
  \otimes \bm B$.  \myitem The $n$-th column of the identity matrix
  $\identitymat_N$ is represented by $\canonicalvec{N}{n}$.  \myitem
  If $\bm A$ is a matrix and $\bm x$ a vector, then $||\bm x||^2_{\bm
    A}\define \bm x\transpose \bm A\inv \bm x$ and $||\bm x
  ||_2\define ||\bm x||_{\identitymat}$.  \myitem  $\pdset^N$
  represents the cone of $N\times N$ positive definite matrices.
  \myitem Finally, $\delta[\cdot]$ stands for the Kronecker delta,
  \myitem and $\expectednb$ for  expectation.

\end{myitemize}

\section{Problem Formulation}
\label{sec:problemdef}

\cmt{Definitions}
\begin{myitemize}
  \myitem \cmt{graph} A time-varying
  graph\footnote{\change{See~\cite{wehmuth2015unifyingconf} and
      references therein for alternative representations of
      time-varying graphs.}  } is a tuple
  $\graph:=(\vertexset,\{\spaceadjacencymat\timenot{\timeind}\}_{\timeind=1}^{\timenum})$,
  where $\vertexset:=\{v_1, \ldots, v_\vertexnum\}$ is the vertex set
  and $\spaceadjacencymat\timenot{\timeind}$ is an
  $\vertexnum\times\vertexnum$ adjacency matrix whose
  $(\vertexind,\vertexindp)$-th entry
  $\spaceadjacencymatentry\timevertexvertexnot{\timeind}{\vertexind}{\vertexindp}$
  assigns a weight to the  pair of vertices $(v_\vertexind,v_\vertexindp)$
  at time $\timeind$. A time-invariant graph is a special case with
  $\spaceadjacencymat\timenot{\timeind}=
  \spaceadjacencymat\timenot{\timeindp}~\forall\timeind,\timeindp$. \change{As
    usual, see
    e.g.~\cite[Ch. 2]{kolaczyck2009},\cite{shuman2013emerging,belkin2006manifold}},
  this paper assumes that $\graph$ (i) has non-negative weights
  ($\spaceadjacencymatentry\timevertexvertexnot{\timeind}{\vertexind}{\vertexindp}\geq
  0~\forall \vertexind,\vertexindp,\timeind$); (ii) no self-links
  ($\spaceadjacencymatentry\timevertexvertexnot{\timeind}{\vertexind}{\vertexind}=0~\forall\vertexind,\timeind$);
  and, (iii) it is undirected
  ($\spaceadjacencymatentry\timevertexvertexnot{\timeind}{\vertexind}{\vertexindp}=
  \spaceadjacencymatentry\timevertexvertexnot{\timeind}{\vertexindp}{\vertexind}~\forall
  \vertexind,\vertexindp,\timeind$).
  \cmt{topology} The edge set is defined as
  $\edgeset\timenot{\timeind} :=\{(v_\vertexind,v_\vertexindp)\in
  \vertexset \times \vertexset:
  \spaceadjacencymatentry\timevertexvertexnot{\timeind}{\vertexind}{\vertexindp}\neq
  0\}$, and two vertices $v$ and $v'$ are said to be \emph{adjacent},
  \emph{connected}, or \emph{neighbors} at time $\timeind$ if
  $(v,v')\in \edgeset\timenot{\timeind}$.


  \myitem \cmt{time series}A time-evolving function or signal on a
  graph,\footnote{The entire framework can naturally be extended to
    accommodate complex-valued functions $\signalfun$.}  is a map $
  \signalfun:\vertexset\times\timeset \rightarrow \mathbb{R}$, where
  $\timeset\define\{1,\ldots,\timenum\}$ is the set of time
  indices. The value $\signalfun(v_\vertexind,\timeind)$ of
  $\signalfun$ at vertex $v_\vertexind$ and time $\timeind$, or its
  shorthand version $\signalfun\timevertexnot{\timeind}{\vertexind}$,
  can be thought of as the value of an attribute of $v_\vertexind\in
  \vertexset$ at time $\timeind$. \cmt{example}In a social network,
  $\signalfun\timevertexnot{\timeind}{\vertexind}$ may denote the
  annual income of person $v_\vertexind$ at year $\timeind$. Function
  values at time $\timeind$ will be collected in
  $\signalvec\timenot{\timeind}:=[\signalfun\timevertexnot{\timeind}{1},\ldots,
    \signalfun\timevertexnot{\timeind}{N}]\transpose$.
\end{myitemize}

 \cmt{observations}
\begin{myitemize}
  \myitem\cmt{scalar form}At time $\timeind$, the vertices with
  indices in the \change{time-dependent and arbitrary}  set
  $\sampleset\timenot{\timeind}:=\{\vertexind\sampletimenot{1}{\timeind},\ldots,\vertexind\sampletimenot{\samplenum\timenot{\timeind}}{\timeind}\}$,
  $1\leq
  \vertexind\sampletimenot{1}{\timeind}<\cdots<\vertexind\sampletimenot{\samplenum\timenot{\timeind}}{\timeind}\leq
  \vertexnum$, are observed. The resulting samples can be expressed as
  $\observationfun\sampletimenot{\sampleind}{\timeind} =
  \signalfun\timevertexnot{\timeind}{\vertexind_\sampleind\timenot{\timeind}}
  + \observationnoisefun\sampletimenot{\sampleind}{\timeind},
  \sampleind=1,\ldots,\samplenum\timenot{\timeind}$, where
  $\observationnoisefun\sampletimenot{\sampleind}{\timeind}$ models
  observation error.  \cmt{intuition}In social networks, this
  encompasses scenarios where a subset of persons have been surveyed
  about the attribute of interest; e.g. their annual income.
  \myitem\cmt{vector-matrix form}By letting
  $\observationvec\timenot{\timeind} \define
  [\observationfun\sampletimenot{1}{\timeind},\ldots,\observationfun\sampletimenot{\samplenum\timenot{\timeind}}{\timeind}]\transpose$,
  the observations can be conveniently expressed as
\begin{equation}
\label{eq:observationsvec}
\observationvec\timenot{\timeind} =\samplemat\timenot{\timeind} \signalvec\timenot{\timeind}  + \observationnoisevec\timenot{\timeind},\quad\timeind=1,\ldots,\timenum
\end{equation}
where $\observationnoisevec\timenot{\timeind} \define
[\observationnoisefun\sampletimenot{1}{\timeind},\ldots,\observationnoisefun\sampletimenot{\samplenum\timenot{\timeind}}{\timeind}]\transpose$,
and the $\samplenum\timenot{\timeind}\times \vertexnum$ sampling
matrix $\samplemat\timenot{\timeind}$ contains ones at positions
$(\sampleind,\vertexind\sampletimenot{\sampleind}{\timeind})$,
$\sampleind=1,\ldots,\samplenum\timenot{\timeind}$ and zeros
elsewhere.

\end{myitemize}

\cmt{Problem formulation}
\begin{myitemize}

\myitem\cmt{goal=reconstruct}The broad goal of this paper is to ``reconstruct''
$\signalfun$ from the observations $\{\observationvec\timenot{\timeind}\}_{\timeind=1}^\timenum$ in \eqref{eq:observationsvec}. Two
 formulations will be considered:
  \myitem\cmt{batch problem formulation}  in the batch 
  formulation, one aims at
\begin{myitemize}
\myitem finding $\{\signalvec\timenot{\timeind}\}_{\timeind=1}^\timenum$
  \myitem given $\graph$, 
  the sample locations
  $\{\samplemat\timenot{\timeind}\}_{\timeind=1}^\timenum$, and all
  observations
  $\{\observationvec\timenot{\timeind}\}_{\timeind=1}^\timenum$.
\end{myitemize}
\myitem\cmt{online problem formulation}In the online formulation,
one is given $\graph$ together with $\samplemat\timenot{\timeind}$ and
$\observationvec\timenot{\timeind}$ at time $\timeind$. The goal is to
find $\signalvec\timenot{\timeind}$, possibly based on a previous estimate of
$\signalvec\timenot{\timeind-1}$, with bounded complexity per time
slot $\timeind$, even if $\timenum\rightarrow \infty$.
\end{myitemize}
\cmt{Prior info=smoothness}To solve these problems, no explicit
parametric model for the temporal or spatial evolution of $\signalfun$
will be adopted. For instance, one can solely rely on the assumption
that $\signalfun$ evolves smoothly over both space and time, yet more
structured dynamics can also be incorporated if known.

\section{Background on kernel-based  reconstruction}
\label{sec:background}

\cmt{Sec. overview}This section reviews the existing framework for
kernel-based reconstruction of time-invariant functions, a special
case of which is the batch problem in Sec.~\ref{sec:problemdef} 
when $\timenum=1$. Reflecting this scenario, the notation will be devoid of time indices.  \cmt{single-snapshot problem
  form.}As a result, the problem becomes finding
$\signalvec\in \rfield^{\vertexnum}$ given
$\spaceadjacencymat\in\rfield_+^{\vertexnum\times\vertexnum}$,
$\samplemat\in\{0,1\}^{\samplenum\times\vertexnum}$, and
$\observationvec =
\samplemat \signalvec +\observationnoisevec\in\rfield^\samplenum$.

\cmt{LS estimator}At first, one may feel tempted to seek a
least-squares estimate $\signalestvec
=\argmin_{\signalvec}||\observationvec-\samplemat\signalvec||_2^2$,
but noting that the $\vertexnum$ unknowns in $\signalvec$ cannot be
generally identified from the $\samplenum\leq \vertexnum$ samples in
$\observationvec$ dismisses such an approach.
\cmt{reg. estimator}This underdeterminacy prompts estimates of the
form $\signalestvec
=\argmin_{\signalvec}||\observationvec-\samplemat\signalvec||_2^2+\regpar\regfun(\signalvec)$,
where $\regpar>0$ and the regularizer $\regfun(\signalvec)$ promotes a
certain structure in $\signalvec$.
\cmt{Laplacian reg.}A customary $\regfun(\signalvec)$ encourages smooth estimates by
penalizing functions that exhibit pronounced variations among
neighboring vertices,
\begin{myitemize}
\myitem for instance by means of  the so-called Laplacian regularizer
\begin{align}
\label{eq:graphvariationdef}
\graphvariation(\signalvec) \define\frac{1}{2} \sum_{\vertexind=1}^\vertexnum
\sum_{\vertexindp=1}^\vertexnum \spaceadjacencymatentry\vertexvertexnot{\vertexind}{\vertexindp}(\signalfun\vertexnot{\vertexind}-\signalfun\vertexnot{\vertexindp})^2
\end{align}
which heavily penalizes differences between function values at
vertices connected by strong links (large
$\spaceadjacencymatentry\vertexvertexnot{\vertexind}{\vertexindp}$). \change{
  Expression \eqref{eq:graphvariationdef} formalizes the notion of
  smoothness introduced in Sec.~\ref{sec:intro}, according to which a
  function is smooth if it takes similar values at neighboring
  vertices. Since $\graphvariation(\signalvec)$ is small if
  $\signalvec$ is smooth, and large otherwise,
  $\graphvariation(\signalvec)$ acts as a proxy quantifying smoothness
  of $\signalvec$, in the sense that given two functions $\signalvec$
  and $\signalvec'$, the former is said to be smoother than the latter
  iff $\graphvariation(\signalvec)<\graphvariation(\signalvec')$ and
  vice versa. More general proxies are reviewed next.}


\myitem \cmt{Laplacian def.}Upon defining the $\vertexnum \times
\vertexnum$ \emph{Laplacian} matrix
$\spacelaplacianmat:=\diag{\spaceadjacencymat\bm
  1}-\spaceadjacencymat$, the functional in
\eqref{eq:graphvariationdef} can be rewritten after some algebra as
$\graphvariation(\signalvec) =
\signalvec\transpose\spacelaplacianmat\signalvec$; see
e.g.~\cite[Ch. 2]{kolaczyck2009}. It readily follows
from~\eqref{eq:graphvariationdef} that
$\graphvariation(\signalvec)\geq 0~\forall \signalvec$, which in turn
implies that $\spacelaplacianmat$ is positive semidefinite. %
\myitem\cmt{Laplacian evd}Therefore, $\spacelaplacianmat$ admits an
eigenvalue decomposition $\spacelaplacianmat=\spacelaplacianevecmat
\diag{\spacelaplacianevalvec}\spacelaplacianevecmat\transpose$, where
\begin{myitemize}
\myitem the eigenvectors in $\spacelaplacianevecmat\define[
\spacelaplacianevec_1,\ldots,\spacelaplacianevec_\vertexnum]$ and the 
\myitem eigenvalues in
$\spacelaplacianevalvec\define[\spacelaplacianevalvertexnot{1},\ldots,\spacelaplacianevalvertexnot{\vertexnum}]$
are sorted so that $0=\spacelaplacianevalvertexnot{1}\leq\ldots\leq\spacelaplacianevalvertexnot{\vertexnum}$.
\end{myitemize}
By letting
$\fouriersignalfun\vertexnot{\vertexind}\define(\spacelaplacianevec_\vertexind)\transpose
\signalvec$, one finds that 
\begin{align}
\label{eq:gvlambda}
\graphvariation(\signalvec) = \sum_{\vertexind=1}^\vertexnum
\spacelaplacianevalvertexnot{\vertexind}
|\fouriersignalfun\vertexnot{\vertexind}|^2
\end{align}
which means that $\graphvariation(\signalvec)$ is the weighted
superposition of the magnitude of the projections of $\signalvec$ onto
the eigenvectors of $\spacelaplacianmat$ with weights given by
the corresponding eigenvalues.
\myitem\cmt{freq. interpretation}As described next, 
\eqref{eq:gvlambda} provides an insightful interpretation of
$\graphvariation(\signalvec)$ in a transformed domain.  Specifically,
a number of works advocate the term \emph{graph Fourier transform} or
frequency representation of $\signalfun$ to refer to
$\{\fouriersignalfun\vertexnot{\vertexind}\}_{\vertexind=1}^\vertexnum$;
see e.g.~\cite{shuman2013emerging}. 
\begin{myitemize}
\myitem\cmt{Fourier tr.}The main argument resides in  that
$\{\spacelaplacianevec_\vertexind\}_{\vertexind=1}^\vertexnum$ play a role
analogous  to complex exponentials in signal processing for time
signals, in the sense that (i) complex exponentials are
eigensignals of the continuous counterpart of the Laplacian operator
$\signalvec \mapsto \spacelaplacianmat\signalvec$, and (ii)
$\{\spacelaplacianevec_\vertexind\}_{\vertexind=1}^\vertexnum$ are 
eigensignals of the so-called linear, shift-invariant 
filters~\cite{sandryhaila2013discrete}, which are the graph
counterparts of linear, time-invariant filters in signal processing for
time signals.  Thus, $\signalvec =
\sum_{\vertexind=1}^\vertexnum\fouriersignalfun\vertexnot{\vertexind}\spacelaplacianevec\vertexnot{\vertexind}$
resembles in some sense the synthesis equation of the Fourier
transform, and one can therefore interpret
$\{\spacelaplacianevec_\vertexind\}_{\vertexind=1}^\vertexnum$ as a
Fourier basis.
\myitem\cmt{Freq. interpr.}Because
$\spacelaplacianevalvertexnot{1}\leq\ldots\leq\spacelaplacianevalvertexnot{\vertexnum}$,
it follows from $\graphvariation(\spacelaplacianevec_\vertexind)
=(\spacelaplacianevec_\vertexind)\transpose \spacelaplacianmat
\spacelaplacianevec_\vertexind=\spacelaplacianevalvertexnot{\vertexind}$
that $\graphvariation(\spacelaplacianevec_1)\leq\ldots\leq
\graphvariation(\spacelaplacianevec_\vertexnum)$. Hence, sorting the
eigenvectors
$\{\spacelaplacianevec_\vertexind\}_{\vertexind=1}^\vertexnum$ in
increasing order of their associated eigenvalue is tantamount to
sorting them in decreasing order of smoothness. Similarly, the complex
exponentials in the traditional Fourier basis are indexed by their
frequency, which can be thought of as an (inverse) proxy of
time-domain smoothness. Comparing both scenarios suggests interpreting
$\spacelaplacianevalvertexnot{\vertexind}$, or the index $\vertexind$,
as the \emph{graph frequency} of $\spacelaplacianevec_{\vertexind}$.

\end{myitemize}

\end{myitemize}

\cmt{generalization\ra Laplacian kernels}Back to~\eqref{eq:gvlambda}, it is seen
that $\graphvariation(\signalvec)$ penalizes high-frequency components
more heavily than low-frequency ones, thus promoting estimates with a
``low-pass'' graph Fourier transform. A finer control of how
energy is distributed across frequency can be attained
upon applying a transformation
$\frequencyweightfun:\rfield\rightarrow\rfield_+$ to
$\spacelaplacianevalvertexnot{\vertexind}$, giving rise to regularizers
of the form
\begin{subequations}
\label{eq:generalizedlaplacianregularizer}
\begin{align}
\label{eq:gvkernel}
\laplaciankernelregfun(\signalvec) = \sum_{\vertexind=1}^\vertexnum
\frequencyweightfun(\spacelaplacianevalvertexnot{\vertexind})
|\fouriersignalfun\vertexnot{\vertexind}|^2
=\signalvec\transpose\fullkernelmat\pinv\signalvec
\end{align}
where 
\begin{align}
\label{eq:laplaciankerneldef}
\fullkernelmat\pinv\define \frequencyweightfun(\spacelaplacianmat)
\define\spacelaplacianevecmat\transpose\diagnb\{\frequencyweightfun(\spacelaplacianevalvec)\}
\spacelaplacianevecmat
\end{align}
\end{subequations}
is referred to as \emph{Laplacian kernel}~\cite{smola2003kernels}.
Table~\ref{tab:spectralweightfuns} summarizes some well-known examples
arising with specific choices of $\frequencyweightfun$.

\begin{table}
\begin{center}
  \begin{tabular}{|p{2.4cm} | c | p{2.2cm}|}
    \hline
\textbf{Kernel name}  & \textbf{Function}  & \textbf{Parameters} \\
    \hline
    \hline
Diffusion kernel~\cite{kondor2002diffusion}     &
$r(\lambda)=\exp\{\sigma^2\lambda/2\}$  & 
$\sigma^2\geq 0$
\\
    \hline
$p$-step random walk kernel~\cite{smola2003kernels}    & $r(\lambda) =
(a-\lambda)^{-p}$ & $a\geq 2$, $p$  positive integer \\
    \hline
Laplacian
  regularization~\cite{smola2003kernels,zhou2004regularization,shuman2013emerging}
  & $r(\lambda)=1 + \sigma^2\lambda$  & $\sigma^2$ sufficiently large\\
    \hline
Bandlimited~\cite{romero2016multikernel}    &  
$\begin{aligned}
\label{eq:defrbl}
\frequencyweightfun(\laplacianeval) =  
\begin{cases}
1/\beta & \laplacianeval\leq\laplacianeval_\text{max}\\
\beta & \text{otherwise}
\end{cases}
\end{aligned}$
 &
 $\beta>0$ sufficiently large,   $\lambda_\text{max}$
 \\
    \hline
  \end{tabular}
\end{center}
\caption{Common spectral weight functions.}
\label{tab:spectralweightfuns}
\end{table}

\cmt{any kernel\ra KRR}Further broadening the scope of the generalized
Laplacian kernel regularizers in
\eqref{eq:generalizedlaplacianregularizer}, the so-called \emph{kernel
  ridge regression} (KRR) estimators are given by
\begin{equation}
\label{eq:krrnotime}
  \signalestvec
  := \argmin_{\rkhsvec} \frac{1}{\samplenum} 
  ||\observationvec - \samplemat\rkhsvec||^2_2
  + \regpar \rkhsvec\transpose \fullkernelmat\pinv \rkhsvec
\end{equation}
for an arbitrary positive semidefinite matrix $\fullkernelmat$, not
necessarily a Laplacian kernel. The user-selected parameter
$\regpar>0$ balances the importance of the regularizer relative to the
fitting term
$\samplenum\inv||\observationvec-\samplemat\signalvec||_2^2$.
\cmt{advocate KRR}KRR estimators have well-documented merits and solid
grounds on statistical learning theory; see e.g.~\cite{scholkopf2002}.
\cmt{more general est.}Different regularizers and fitting functions
lead to even more general algorithms; see
e.g.~\cite{romero2016multikernel}.

\section{Reconstruction of time series on graphs}
\label{sec:reconstruction}

\cmt{sec. summ}The framework in Sec.~\ref{sec:background} cannot
accommodate functions evolving over both space and time. This section
generalizes this framework through the notion of \emph{graph
  extension} to flexibly exploit spatial and temporal dynamics.


\cmt{time agnostic}An immediate approach to reconstructing
time-evolving functions is to apply \eqref{eq:krrnotime} separately
for each $\timeind=1,\ldots,\timenum$, yielding the instantaneous
estimator (IE)
\begin{align}
\label{eq:timeagnostic}
  \signalestvec_\text{IE}\timenot{\timeind}
  :=& \argmin_{\rkhsvec} \frac{1}{\samplenum\timenot{\timeind}} 
  ||\observationvec\timenot{\timeind} - \samplemat\timenot{\timeind}\rkhsvec||^2_2
  + \regpar \rkhsvec\transpose \fullkernelmat\pinv\timenot{\timeind}
  \rkhsvec.
\end{align}
\cmt{critic=no time info}Unfortunately, this estimator does not
account for the possible relation between e.g. 
$\signalfun\timevertexnot{\timeind}{\vertexind}$ and
$\signalfun\timevertexnot{\timeind-1}{\vertexind}$. If, for instance,
$\signalfun$ varies slowly over time, an estimate of
$\signalfun\timevertexnot{\timeind}{\vertexind}$ may as well benefit
from leveraging observations
$\observationfun\sampletimenot{\sampleind}{\timeindaux}$ at time
instants
$\timeindaux\neq\timeind$. Exploiting
temporal dynamics potentially reduces the number of sampled vertices
required to attain a target estimation performance, which in turn can markedly
reduce sampling costs.

\cmt{Graph extension}
\begin{myitemize}
  \myitem\cmt{Extended graph}Incorporating temporal dynamics into
  kernel-based reconstruction, which can only handle a single snapshot
  (cf.  Sec.~\ref{sec:background}), necessitates an appropriate
  reformulation of time-evolving function reconstruction as a problem
  of reconstructing a time-invariant function. An appealing
  possibility is to replace $\graph$ with its \emph{extended version}
  $\extendedgraph\define( \extendedvertexset,\extendedadjacencymat)$,
  where each vertex in $\vertexset$ is replicated $\timenum$ times to
  yield the extended vertex set
  $\extendedvertexset\define\{v\timevertexnot{\timeind}{\vertexind},~\vertexind=1,\ldots,\vertexnum,~\timeind=1,\ldots,\timenum\}$,
  and \myitem\cmt{extended adjacency}the
  $(\vertexind+\vertexnum(\timeind-1),\vertexindp+\vertexnum(\timeindp-1))$-th
  entry of the $\timenum\vertexnum\times\timenum\vertexnum$ extended
  adjacency matrix $\extendedadjacencymat$ equals the weight of the
  edge
  $(v\timevertexnot{\timeind}{\vertexind},v\timevertexnot{\timeindp}{\vertexindp})$. The
  time-varying function $\signalfun$ can thus be replaced with its extended
  time-invariant counterpart
  $\extendedsignalfun:\extendedvertexset\rightarrow\rfield$
with  $\extendedsignalfun(v\timevertexnot{\timeind}{\vertexind})=\signalfun\timevertexnot{\timeind}{\vertexind}$.

\myitem\cmt{focus on extended graphs where}
\begin{myitemize}
\myitem\cmt{spatial connect. is given}This paper focuses on graph
extensions respecting the connectivity of $\graph$ per time slot
$\timeind$, that is,
$\{v\timevertexnot{\timeind}{\vertexind}\}_{\vertexind=1}^\vertexnum$
are connected according to $\spaceadjacencymat\timenot{\timeind}$,
$\forall \timeind$:
\myitem\cmt{summary}
\change{
  \begin{mydefinitionhere}
    Let   $\vertexset:=\{v_1, \ldots,
    v_\vertexnum\}$ denote a vertex set and let
    $\graph:=(\vertexset,\{\spaceadjacencymat\timenot{\timeind}\}_{\timeind=1}^{\timenum})$
    be a time-varying graph. A graph $\extendedgraph$ with vertex set
    $\extendedvertexset\define\{v\timevertexnot{\timeind}{\vertexind},~\vertexind=1,\ldots,\vertexnum,~\timeind=1,\ldots,\timenum\}$
    and $\vertexnum\timenum\times\vertexnum\timenum$ adjacency matrix
    $\extendedadjacencymat$ is an extended graph of $\graph$ if the
    $\timeind$-th $\vertexnum\times\vertexnum$ diagonal block of
    $\extendedadjacencymat$ equals
    $\spaceadjacencymat\timenot{\timeind}$.
  \end{mydefinitionhere}
  In general, there exist multiple graph extensions for a given
  time-varying graph. This is because only the diagonal blocks of
  $\extendedadjacencymat$ are dictated by
  $\{\spaceadjacencymat\timenot{\timeind}\}_{\timeind=1}^{\timenum}$,
  \myitem\cmt{time connect. freely selected}whereas the remaining
  entries of $\extendedadjacencymat$ can be freely selected. In the
  reconstruction problem, one is interested in selecting such off-diagonal
  entries to capture the space-time dynamics of $\signalfun$.  }
\myitem\cmt{example=block tridiag.}As an example, consider an extended
graph with
\begin{align}
\label{eq:timevaryingextendedadjacencymat}
&  \extendedadjacencymat=
 \tridiag\{\spaceadjacencymat\timenot{1},\ldots,\spaceadjacencymat\timenot{\timenum};\timeconnectionmat\timenot{2},\ldots,\timeconnectionmat\timenot{\timenum}\}
\end{align}
where
$\timeconnectionmat\timenot{\timeind}\in\rfield_+^{\vertexnum\times\vertexnum}$
connects 
$\{v\timevertexnot{\timeind-1}{\vertexind}\}_{\vertexind=1}^\vertexnum$
to
$\{v\timevertexnot{\timeind}{\vertexind}\}_{\vertexind=1}^\vertexnum$,
$\timeind=2,\ldots,\timenum$. For instance, one can connect each vertex
to its neighbors at the previous time instant by setting
$\timeconnectionmat\timenot{\timeind}=\spaceadjacencymat\timenot{\timeind-1}$,
or one can connect each vertex to its replicas at adjacent time
instants by setting $\timeconnectionmat\timenot{\timeind}$ to be
diagonal. Fig.~\ref{fig:extendedgraph} pictorially illustrates the latter
choice.
\end{myitemize}

\begin{figure}
\centering
   \includegraphics[width=.4\textwidth]{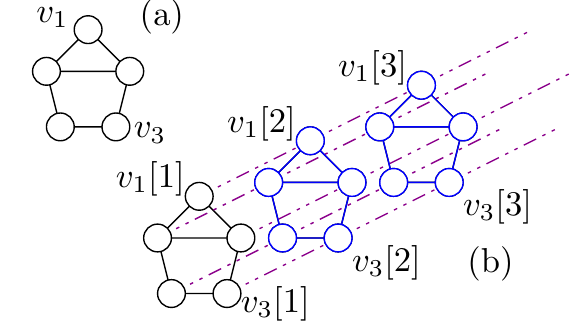}
\caption{(a) Original graph $\graph$. (b) Extended graph
  $\extendedgraph$ for
  diagonal~$\timeconnectionmat\timenot{\timeind}$. \change{Edges connecting
  vertices at the same time instant are represented by solid lines
  whereas edges connecting vertices at different time instants are
  represented by dot-dashed lines.}
}\label{fig:extendedgraph}
\end{figure}

\myitem\cmt{$\timeind$ \ra becomes spatial dimension}Notice that the extended
graph treats the time dimension just as the spatial dimension. Thus,
the flexibility of graphs to convey relational information carries
over to the time domain. As a major benefit,  this approach lays the
grounds for the design of doubly-selective kernels in
Sec.~\ref{sec:titopologies}. 
\end{myitemize}
\cmt{Estimation}The extended graph also enables a generalization of
the estimators in Sec.~\ref{sec:background} to reconstruct
time-evolving functions. The rest of this section develops two KRR
estimators along these lines.

\begin{myitemize}
\myitem\cmt{Proposed batch criterion}Consider first the \emph{batch
  formulation}, where all the
$\extendedsamplenum\define\sum_{\timeind=1}^\timenum\samplenum
\timenot{\timeind}$ samples in $\extendedobservationvec\define [
  \observationvec\transpose\timenot{1},\ldots,\observationvec\transpose\timenot{\timenum}
]\transpose$ are available, and the goal is to estimate
$\extendedrkhsvec\define[\rkhsvec\transpose\timenot{1},\ldots,\rkhsvec\transpose\timenot{\timenum}]\transpose$. Directly
applying the KRR criterion in~\eqref{eq:krrnotime} to reconstruct
$\extendedsignalvec$ on the extended graph $\extendedgraph$ yields
\begin{subequations}
\begin{equation}
\label{eq:batchcriterion}
  \extendedsignalestvec
  := \argmin_{\extendedrkhsvec} 
  ||\extendedobservationvec - \extendedsamplemat\extendedrkhsvec||^2_{\samplenumdiagmat}
  + \regpar \extendedrkhsvec\transpose \extendedfullkernelmat\pinv \extendedrkhsvec
\end{equation}
where
\begin{myitemize}
\myitem  $ \extendedfullkernelmat$ is now a
$\timenum\vertexnum\times\timenum\vertexnum$ ``space-time'' kernel
matrix to be designed in Sec.~\ref{sec:kernels}, 
\myitem
$\extendedsamplemat\define\bdiag{\samplemat\timenot{1},\ldots,\samplemat\timenot{\timenum}
}$, and
 \myitem  $\samplenumdiagmat \define \bdiag{
\samplenum\timenot{1}\bm I_{\samplenum\timenot{1}}
,
\ldots,
\samplenum\timenot{\timenum}\bm I_{\samplenum\timenot{\timenum}}
}$.
\end{myitemize}
\cmt{closed form}If 
$\extendedfullkernelmat$ is invertible, \eqref{eq:batchcriterion} can be solved in
closed form as
\begin{align}
  \extendedsignalestvec=
\extendedfullkernelmat
\extendedsamplemat\transpose
(
\extendedsamplemat
\extendedfullkernelmat
\extendedsamplemat\transpose
+
\regpar \samplenumdiagmat
)\inv
\extendedobservationvec.
\label{eq:closedbatchform}
\end{align}
\end{subequations}
\cmt{when it reduces to t agnostic}\change{For the special
  $\extendedfullkernelmat\pinv =
  \bdiag{\fullkernelmat\pinv\timenot{1},\ldots,\fullkernelmat\pinv\timenot{\timenum}}$,
  where $\fullkernelmat\timenot{\timeind}$ is an
  $\vertexnum\times\vertexnum$ kernel matrix for $\graph$ at time
  $\timeind$, then \eqref{eq:batchcriterion} separates into $\timenum$
  sub-problems, each as in \eqref{eq:timeagnostic}.  This implies that
  only matrices  $\extendedfullkernelmat\pinv$ with non-zero entries
  off its block diagonal are capable of  accounting for 
  temporal dynamics.}

\end{myitemize}

\label{sec:onlineestimator}

\begin{myitemize}
\myitem\cmt{Proposed online criterion}In the \emph{online formulation}, one aims to estimate 
$\signalvec\timenot{\timeind}$ after the
$\extendedsamplenum\timenot{\timeind}\define\sum_{\timeindaux=1}^\timeind\samplenum
\timenot{\timeindaux}$ samples in
$\extendedobservationvec\timenot{\timeind}\define [
  \observationvec\transpose\timenot{1},\ldots,\observationvec\transpose\timenot{\timeind}
]\transpose$ become available. Based on these samples, the KRR estimate of
$\extendedsignalvec$, denoted as $\extendedsignalestvec\giventimenot{\timeind}$, is clearly
\begin{subequations}
\label{eq:onlinecriteriongeneral}
	\begin{eqnarray}
  \extendedsignalestvec\giventimenot{\timeind}
  :=& \argmin_{\extendedrkhsvec} 
  ||\extendedobservationvec\timenot{\timeind} -
  \extendedsamplemat\timenot{\timeind}\extendedrkhsvec||^2_{
\samplenumdiagmat\timenot{\timeind}
}
  + \regpar \extendedrkhsvec\transpose
  \extendedfullkernelmat\inv
\extendedrkhsvec
\label{eq:onlinecriterion}\\
  =&
\extendedfullkernelmat
\extendedsamplemat\transpose\timenot{\timeind}
(
\extendedsamplemat\timenot{\timeind}
\extendedfullkernelmat
\extendedsamplemat\transpose\timenot{\timeind}
+
\regpar \samplenumdiagmat\timenot{\timeind}
)\inv
\extendedobservationvec\timenot{\timeind}.
\label{eq:onlinecriterionsol}
\end{eqnarray}
\end{subequations}
where
\begin{myitemize}\myitem{}$\extendedfullkernelmat$ is assumed invertible for simplicity,
\myitem$\samplenumdiagmat\timenot{\timeind}\define \bdiag{
  \samplenum\timenot{1}\bm I_{\samplenum\timenot{1}} , \ldots,
  \samplenum\timenot{\timeind}\bm I_{\samplenum\timenot{\timeind}} }$,
\myitem and
$\extendedsamplemat\timenot{\timeind}\define[\diag{\samplemat\timenot{1},\ldots,\samplemat\timenot{\timeind}},\bm0_{
    \extendedsamplenum\timenot{\timeind} \times
    (\timenum-\timeind)\vertexnum }]
\in\{0,1\}^{\extendedsamplenum\timenot{\timeind}\times
  \timenum\vertexnum}$.
\end{myitemize}
\change{ The estimate in \eqref{eq:onlinecriteriongeneral} comprises
  the per slot estimates
  $\{\signalestvec\timegiventimenot{\tau}{\timeind}\}_{\tau=1}^\timenum$;
  that is, $ \extendedsignalestvec\giventimenot{\timeind} \define[
    \signalestvec\transpose\timegiventimenot{1}{\timeind},
    \signalestvec\transpose\timegiventimenot{2}{\timeind}, \ldots,
    \signalestvec\transpose\timegiventimenot{\timenum}{\timeind}
  ]\transpose $ with $
  \signalestvec\timegiventimenot{\tau}{\timeind}=[
    \signalestfun\timegiventimevertexnot{\tau}{\timeind}{1}, \ldots,
    \signalestfun\timegiventimevertexnot{\tau}{\timeind}{\vertexnum}]\transpose$,
  where $\signalestvec\timegiventimenot{\tau}{\timeind}$ (respectively
  $ \signalestfun\timegiventimevertexnot{\tau}{\timeind}{\vertexind}$)
  is the KRR estimate of $\signalvec\timenot{\tau}$
  ($\signalfun\timevertexnot{\tau}{\vertexind}$) given the
  observations up to time $\timeind$. Observe that, with this notation, it follows
  that
\begin{align}
\label{eq:onlinecriteriontimet}
\signalestvec\timegiventimenot{\tau}{\timeind} &=
(\canonicalvec{\timenum}{\tau}\transpose\otimes
\identitymat_{\vertexnum})
\extendedsignalestvec\giventimenot{\timeind}
\end{align}
for all $\timeind,\tau$.   }

\cmt{online algo}\change{Regarding $\timeind$ as the present,
  \eqref{eq:onlinecriteriongeneral} therefore provides estimates of
  past, present, and future values of $f$. The solution to the online
  problem formulated in Sec.~\ref{sec:problemdef} includes the
  sequence of present KRR estimates for all $t$, that is,
  $\{\signalestvec\timegiventimenot{\timeind}{\timeind}\}_{\timeind=1}^\timenum$.
  This can be obtained by solving \eqref{eq:onlinecriterion} in closed
  form per $\timeind$ as in \eqref{eq:onlinecriterionsol} and then
  applying \eqref{eq:onlinecriteriontimet}. However, such an approach
  does not yield a desirable online algorithm since its complexity per
  time slot is cubic in $\timeind$ (see
  Remark~\ref{remark:complexity}) and therefore increasing with $t$.
  For this reason, this approach is not satisfactory since the online
  problem formulation in Sec.~\ref{sec:problemdef} requires the
  complexity per time slot of the desired algorithm to be bounded.  An
  algorithm that does satisfy this bounded-complexity requirement yet
  provides the exact KRR estimate is developed next for the case where
  the kernel matrix is any positive definite  matrix $
  \extendedfullkernelmat$ satisfying}
\begin{align}
\label{eq:invblocktriagonal}
 \extendedfullkernelmat\inv&=
\tridiag\{\kernelinvondiagonalmat\timenot{1},\ldots,\kernelinvondiagonalmat\timenot{\timenum}
;\kernelinvoffdiagonalmat\timenot{2},\ldots,\kernelinvoffdiagonalmat\timenot{\timenum}
\}
\end{align}
for some $\vertexnum\times\vertexnum$ matrices
$\{\kernelinvondiagonalmat\timenot{\timeind}\}_{\timeind=1}^\timenum$
and
$\{\kernelinvoffdiagonalmat\timenot{\timeind}\}_{\timeind=2}^\timenum$.
Kernels in this important family are designed in
Sec.~\ref{sec:kernels}.  \change{Broader classes of kernels can be
  accommodated as described in Remark~\ref{remark:nontridiagonal}.
}


\change{The process of developing the desired online algorithm
  involves two steps. The first step expresses
  \eqref{eq:onlinecriterion} as a weighted least-squares problem
  amenable to a KF solver. In the second step, the KF is applied to
  solve such a problem. The first step is accomplished by the  following result.}
\begin{mylemmahere}\label{lem:kf-krreq} 
{For
      $\extendedfullkernelmat$ of the form
      \eqref{eq:invblocktriagonal}, the KRR criterion in
      \eqref{eq:onlinecriterion} boils down to the following
      regularized weighted least-squares objective }
                      \begin{align} 
      & \extendedsignalestvec\giventimenot{\timeind}= 
       \underset{\{\rkhsvec[\tau]\}_{\tau=1}^\timenum }{\argmin}
      ~\sum_{\tau=1}^{\timeind}\frac{1}{\observationnoisevar\timenot{\tau}}||\observationvec\timenot{\tau}
      - \samplemat\timenot{\tau}\rkhsvec\timenot{\tau}
      ||^2\nonumber\\
      &+\sum_{\tau=2}^{\timenum}||\rkhsvec\timenot{\tau}-
      ~\transitionmat\timenot{\tau}\rkhsvec\timenot{\tau-1}||^2_{\plantnoisekernelmat\timenot{\tau}}
      +\rkhsvec\transpose\timenot{1}\plantnoisekernelmat\inv\timenot{1} \rkhsvec\timenot{1}. 
      \label{eq:kfobj} 
      \end{align}
%
  \end{mylemmahere}

 \begin{IEEEproof}
See Appendix~\ref{proof:lemma}.
\end{IEEEproof}

\change{
      Relative to \eqref{eq:onlinecriterion}, matrices $\{\kernelinvondiagonalmat\timenot{\tau},
      \kernelinvoffdiagonalmat\timenot{\tau}\}$ in $\extendedfullkernelmat\inv$ have been replaced in
      \eqref{eq:kfobj} with matrices $\{\plantnoisekernelmat\timenot{\tau},
      \transitionmat\timenot{\tau}\}$, which can be found through
      Algorithm~\ref{algo:rec}.
      }

\change{Although no probabilistic assumption is required throughout
  the derivation of the proposed online algorithm, exploring the link
  between \eqref{eq:kfobj} and the conventional probabilistic setup
  for state estimation provides the intuition behind why
  \eqref{eq:kfobj} can be solved through Kalman filtering. To this
  end, suppose that $\signalvec\timenot{\tau}$ obeys the random model
  $\signalvec\timenot{\tau}=\transitionmat\timenot{\tau}
  \signalvec\timenot{\tau-1} + \plantnoisevec\timenot{\tau}$ for
  $\tau=2, \ldots, \timenum$, initialized by
  $\signalvec\timenot{1}=\plantnoisevec\timenot{1}$, with zero-mean
  noise $\plantnoisevec\timenot{\tau}$ having covariance
  $\plantnoisekernelmat\timenot{\tau}$, and the that the observations
  follow the model $\observationvec\timenot{\tau}=
  \samplemat\timenot{\timeind}\rkhsvec\timenot{\tau}
  +\observationnoisevec\timenot{\tau}$ for $\tau=1, \ldots, \timenum$,
  with $\observationnoisevec\timenot{\tau}$ zero-mean noise having
  covariance $\observationnoisevar\timenot{\tau}\identitymat$.  In
  this state estimation problem, $\transitionmat\timenot{\tau}$ is
  referred to as the state-transition matrix. In this scenario, one
  can easily see that obtaining the maximum a posteriori (MAP) and the
  minimum mean square error (MMSE) estimators  of
  $\extendedsignalestvec$ given the observations up to time $\timenum$
  when
  $\{\plantnoisevec\timenot{\tau},\observationnoisevec\timenot{\tau}\}_{\tau=1}^\timenum$
  are Gaussian distributed reduces to minimizing~\eqref{eq:kfobj}.
  This link suggests that \eqref{eq:kfobj} can be minimized using the
  celebrated KF~\cite[Ch. 17]{strang1997linear}.}

\change{The following result formalizes the latter claim. The resulting
  algorithm, termed KKF, is summarized as
  Algorithm~\ref{algo:kalmanfilter}.  In the probabilistic KF
  terminology, step~\ref{step:prediction} yields the prediction of
  $\rkhsvec\timenot{\timeind}$, step~\ref{step:predictionerror}
  provides the covariance matrix of the prediction error,
  step~\ref{step:gain} yields the Kalman gain,
  step~\ref{step:correction} returns the posterior estimate upon
  correcting the prediction with the innovations scaled by the Kalman
  gain, and step~\ref{step:correctionerror} finds the error of this
  posterior estimate.}
\change{
\begin{mytheoremhere}
\label{prop:KFanytridiag}{For $\extendedfullkernelmat$ of the form
\eqref{eq:invblocktriagonal}, the KKF Algorithm~\ref{algo:kalmanfilter} returns the sequence 
$\{\signalestvec\timegiventimenot{\timeind}{\timeind}\}_{\timeind=1}^\timenum$,
  where $\signalestvec\timegiventimenot{\timeind}{\timeind}$ is given
  by~\eqref{eq:onlinecriteriontimet}.  }
\end{mytheoremhere}
\begin{IEEEproof}
See Appendix~\ref{proof:theorem}.
\end{IEEEproof}
}

\begin{algorithm}[t]
  \caption{Recursion to set parameters of KKF}\label{algo:rec}
 \textbf{Input:} $\kernelinvondiagonalmat\timenot{\timeind}$, $\timeind =
  1,\ldots,\timenum$, 
  $\kernelinvoffdiagonalmat\timenot{\timeind}$, $\timeind =
  2,\ldots,\timenum$.\\
  \begin{algorithmic}[1]
\STATE  \textbf{Set}  $\plantnoisekernelmat\inv\timenot{\timenum}=\kernelinvondiagonalmat\timenot{\timenum}$\\

\STATE \textbf{for} {$\timeind = \timenum,~\timenum-1,\ldots,2$} \textbf{do}
\STATE\quad    $\transitionmat\timenot{\timeind}=-\plantnoisekernelmat\timenot{\timeind}\kernelinvoffdiagonalmat\timenot{\timeind}$\\
\STATE\quad    $\plantnoisekernelmat\inv\timenot{\timeind-1}=\kernelinvondiagonalmat\timenot{\timeind-1}-\transitionmat\transpose\timenot{\timeind}
    \plantnoisekernelmat\inv\timenot{\timeind}\transitionmat\timenot{\timeind}$\\
\end{algorithmic}
  \textbf{Output: } $\plantnoisekernelmat\timenot{\timeind}$, $\timeind =
  1,\ldots,\timenum$, 
  $\transitionmat\timenot{\timeind}$, $\timeind =
  2,\ldots,\timenum$\\
\end{algorithm}

\begin{algorithm}[t]                
 	\caption{Kernel Kalman filter (KKF)}
 	\label{algo:kalmanfilter}    
 	\begin{minipage}{20cm}
  \textbf{Input:} $\{\plantnoisekernelmat\timenot{\timeind}\in\pdset^{\vertexnum}\}_{\timeind =
  1}^\timenum$,  $\{\transitionmat\timenot{\timeind}\in \rfield^{\vertexnum\times\vertexnum}\}_{\timeind =
  2}^\timenum$, \\
\indent\hspace{1cm}$\{\observationvec\timenot{\timeind}\in\rfield^{\samplenum\timenot{\timeind}}\}_{\timeind=1}^\timenum$,
$\{\samplemat\timenot{\timeind}\in\{0,1\}^{\samplenum\timenot{\timeind}\times\vertexnum}\}_{\timeind=1}^\timenum$, \\
\indent\hspace{1cm}$\{\observationnoisevar\timenot{\timeind}>0\}_{\timeind=1}^\timenum$.
\begin{algorithmic}[1]
\STATE\textbf{Set} $\signalestvec\timenot{0|0}=\bm 0$,
                        $\errormat\timegiventimenot{0}{0}=\bm0$,
                        $\transitionmat\timenot{1} = \bm0$

\STATE                    \textbf{for}~{$t=1,\ldots,\timenum$}{  }\textbf{do}\\
\STATE\label{step:prediction}\quad                          $\signalestvec\timegiventimenot{\timeind}{\timeind-1}=\transitionmat\timenot{\timeind}
                          \signalestvec\timegiventimenot{\timeind-1}{\timeind-1}
                          $\\
\STATE\label{step:predictionerror}\quad                          $
                          \errormat\timegiventimenot{t}{t-1}=\transitionmat\timenot{\timeind}\errormat\timegiventimenot{t-1}{t-1}\transitionmat\transpose\timenot{\timeind}+\plantnoisekernelmat\timenot{t}$\\
\STATE\label{step:gain}\quad                          $
  \kalmangainmat\timenot{t}
  =\errormat\timegiventimenot{t}{t-1}\samplemat\transpose\timenot{t}\times$\\
 \hspace{1.5cm} $\times(\observationnoisevar\timenot{\timeind}\identitymat 
+\samplemat\timenot{t}\errormat\timegiventimenot{t}{t-1}\samplemat\transpose\timenot{t})^{-1}$\\
\STATE\label{step:correction}\quad
$  \signalestvec\timenot{t|t}=\signalestvec\timenot{t|t-1}+\kalmangainmat\timenot{t}(\observationvec\timenot{t}-\samplemat\timenot{t}\signalestvec\timenot{t|t-1})
$\\
\STATE\label{step:correctionerror}\quad
$  \errormat\timegiventimenot{t}{t}=(\identitymat-\kalmangainmat\timenot{t}\samplemat\timenot{t})\errormat\timegiventimenot{t}{t-1}$
\end{algorithmic}
  
\textbf{Output:} $\signalestvec\timegiventimenot{\timeind}{\timeind}$, $\timeind =
  1,\ldots,\timenum$;
  $\errormat\timenot{\timeind}$, $\timeind =
  1,\ldots,\timenum$.

 	\end{minipage}
\end{algorithm}


\change{Recapitulating, given $ \extendedfullkernelmat\inv$ in
  \eqref{eq:invblocktriagonal}, one just has to run
  Algorithms~\ref{algo:rec} and~\ref{algo:kalmanfilter} to find the
  online KRR estimate of $\signalfun$ given by
  \eqref{eq:onlinecriteriontimet}. Since the proposed KKF is derived
  within a fully deterministic framework, notions such as mean,
  covariance, statistical independence, or mean-square error are not
  required, yet they have been used to describe the connection with
  the classical KF.  Furthermore, the proposed KKF does not explicitly
  involve any state-space model, which is a major novelty and indeed a
  surprising result of the present paper.  }

\cmt{summary}\change{
}

\cmt{Kernel generalizes probabilistic}\change{The proposed KKF
  generalizes the probabilistic KF since the latter is recovered upon
  setting $\extendedfullkernelmat$ to be the covariance matrix of
  $\extendedrkhsvec$ in the previously mentioned probabilistic
  setup. It is therefore natural that the assumptions required by the
  probabilistic KF are stronger than those involved in the
  KKF. Specifically, in the probabilistic KF,
  $\rkhsvec\timenot{\timeind}$ must adhere to a linear state-space
  model with known transition matrix
  $\transitionmat\timenot{\timeind}$, where the state noise
  $\plantnoisevec\timenot{\timeind}$ is uncorrelated over time and has
  known covariance matrix $\plantnoisekernelmat\timenot{\timeind}$,
  and the observation noise $\observationvec\timenot{\timeind}$ must
  be uncorrelated over time and have known covariance
  matrix. Correspondingly, the performance guarantees of the
  probabilistic KF are also stronger: the resulting estimate is
  optimal in the mean-square error sense among all linear
  estimators. Furthermore, if $\plantnoisevec\timenot{\timeind}$ and
  $\observationvec\timenot{\timeind}$ are jointly Gaussian,
  $\timeind=1,\ldots,\timenum$, then the probabilistic KF estimate is
  optimal in the mean-square error sense among all (not necessarily
  linear) estimators.}  \change{In contrast, the requirements of the
  proposed KKF are much weaker since it only requires $\signalfun$ to
  evolve smoothly with respect to a given extended graph, but the
  guarantees are also weaker; see
  e.g.~\cite[Ch. 5]{scholkopf2002}. However, since the KKF generalizes
  the probabilistic KF, the reconstruction performance of the former
  for judiciously selected $\extendedfullkernelmat$ cannot be worse
  than the reconstruction performance of the latter for any given
  criterion. The caveat, however, is that such a selection is not
  necessarily~easy. }

\begin{myremarkhere}
\label{remark:complexity}
  \cmt{complxt}Algorithm~\ref{algo:kalmanfilter}
requires $\mathcal{O}(\vertexnum^3)$ operations per time slot, whereas
the complexity of evaluating \eqref{eq:onlinecriterionsol} for the
$\timeind$-th time slot is
$\mathcal{O}(\extendedsamplenum^3\timenot{\timeind})$, which increases
with $\timeind$ and becomes eventually prohibitive.  For large
$\timeind$, Algorithm~\ref{algo:kalmanfilter} is computationally more efficient
than a \emph{single} plain evaluation of
\eqref{eq:onlinecriterionsol}: whereas the overall complexity of the
former is $\mathcal{O}(\timeind \vertexnum^3)$, the latter is
$\mathcal{O}(\vertexnum\timenum\extendedsamplenum^2\timenot{\timeind})$, which e.g. for
constant $\samplenum\timenot{\timeind}=\samplenum$ is
$\mathcal{O}(\vertexnum\timenum\timeind^2\samplenum^2)$. 
\end{myremarkhere}
\begin{myremarkhere}\cmt{prediction}Algorithm~\ref{algo:kalmanfilter}
provides estimates of the form
$\signalestvec\timegiventimenot{\timeind}{\timeind}$ and
$\signalestvec\timegiventimenot{\timeind}{\timeind-1}$. To obtain
estimates $\signalestvec\timegiventimenot{\timeind}{\timeindp}$ for
$\timeind>\timeindp+1$, one may set
$\sampleset\timenot{\timeindaux}=\emptyset$ for
$\timeindaux>\timeindp+1$ and execute
Algorithm~\ref{algo:kalmanfilter} up to time
$\timeind$. \cmt{smoothing}Conversely, to obtain estimates
$\signalestvec\timegiventimenot{\timeind}{\timeindp}$ for which
$\timeind<\timeindp$, one may extend Algorithm~\ref{algo:kalmanfilter}
by capitalizing on the notion of Kalman
smoothing~\cite{rauch1965dynamic}.
\end{myremarkhere}
\change{
  \begin{myremarkhere}
\label{remark:nontridiagonal}
    \cmt{non-tridiagonal}Similar to the
  probabilistic KF, which requires  the inverse covariance matrix
  of $\extendedsignalvec$ to be block tridiagonal, the proposed KKF
  requires the inverse kernel matrix to be of the form
  \eqref{eq:invblocktriagonal}. Fortunately, it is straightforward to
  extend both algorithms to accommodate inverse covariance or kernel
  matrices with any number of non-zero diagonals at the price of
  increasing the time interval between consecutive estimates. To
  illustrate such an approach, suppose that $ \extendedfullkernelmat\inv$
  is not block tridiagonal when blocks are of size
  $\vertexnum\times\vertexnum$, but it is block tridiagonal if blocks
  are of size $2\vertexnum\times 2\vertexnum$. In such a case, one can
  use the proposed KKF to estimate
  $\{\signalvec'\timenot{\timeindp}\}_{\timeindp=1}^{\timenum/2}$,
  where $\signalvec'\timenot{\timeindp}\define
  [\signalvec\transpose\timenot{2\timeindp-1},
    \signalvec\transpose\timenot{2\timeindp}]\transpose \in
  \rfield^{2\vertexnum}$, just by replacing
  $\observationvec\timenot{\timeind}$ with
  $\observationvec'\timenot{\timeindp}\define[\observationvec\transpose\timenot{2\timeindp-1},
    \observationvec\transpose\timenot{2\timeindp}]\transpose$,
  $\samplemat\timenot{\timeind}$ with
  $\samplemat'\timenot{\timeindp}\define\bdiag{
    \samplemat\timenot{2\timeindp-1} \samplemat\timenot{2\timeindp}}$,
  and $\observationnoisevec\timenot{\timeind}$ with
  $\observationnoisevec'\timenot{\timeindp}\define
  [\observationnoisevec\transpose\timenot{2\timeindp-1},
    \observationnoisevec\transpose\timenot{2\timeindp}]\transpose$,
  $\timeindp=1,\ldots,\timenum/2$. Note that the sampling interval
  associated with the index
  $\timeindp$ is twice that associated with $\timeind$.
\end{myremarkhere}
}


\end{myitemize}

\section{Design of space-time kernels}
\label{sec:kernels}

\cmt{so far}Sec.~\ref{sec:reconstruction} assumed that the kernel
matrix $\extendedfullkernelmat$ is given and described no methodology
to address its design.  \cmt{Immediate approach=raw Laplacian
  kernel}An immediate approach is to mimic the Laplacian kernels of
Sec.~\ref{sec:background} by setting $\extendedfullkernelmat =
\frequencyweightfun\pinv(\extendedlaplacianmat)$, where
$\extendedlaplacianmat \define
\diagnb\{\extendedadjacencymat\bm1\}-\extendedadjacencymat$ denotes
the Laplacian matrix of the extended graph. Unfortunately, such a
design prevents separate control of the spatial and temporal
variability of the estimates, thus limiting the user's ability to
flexibly account for spatial and temporal information. For instance,
sampling intervals that are small relative to the time dynamics of
$\signalfun$, meaning that $\signalfun$ does not vary significantly
between samples $\timeind-1$ and $\timeind$, favors estimates that
sacrifice spatial smoothness to increase temporal smoothness.

\cmt{Sec. overview}This section proposes families of {space-time
  kernels} for which temporal and spatial smoothness can be separately
tuned.  Sec.~\ref{sec:titopologies} describes designs for
time-invariant topologies, whereas Sec.~\ref{sec:tvtopologies} deals
with the time-varying case.

\subsection{Doubly-selective space-time  kernels}
\label{sec:titopologies}

\cmt{motivation=design in 2d freq. domain}In
Sec.~\ref{sec:background}, the frequency interpretation
of~\eqref{eq:gvlambda} proved decisive to interpret and design
Laplacian kernels for reconstructing time-invariant
functions. Introducing the time dimension in
Sec.~\ref{sec:reconstruction} prompts an analogous methodology, where
kernels are specified in a bidimensional plane of spatio-temporal
frequency; see~\cite{isufi2016graphtemporal} for graph filter design
in this domain.
\cmt{Sec. overview}This section accomplishes this task by generalizing
the Laplacian kernels from Sec.~\ref{sec:background}. How much the
regularizers $\regfun(\extendedrkhsvec)= \extendedrkhsvec\transpose
\extendedfullkernelmat\pinv \extendedrkhsvec$ associated with the
proposed kernels weight each spatial and temporal frequency
component of $\extendedrkhsvec$ can be separately prescribed.
\cmt{this Sec.\ra time-invariant topologies}Throughout this section, a
time-invariant topology will be assumed, i.e.,
$\spaceadjacencymat\timenot{\timeind} =
\spaceadjacencymat,~\timeind=1,\ldots,\timenum$.

\cmt{extend time-inv framework}
\begin{myitemize}
\myitem\cmt{time-inv}Clearly, \eqref{eq:gvkernel} can be rewritten as
$\laplaciankernelregfun(\signalvec)=\frequencyweightvec\transpose(\fouriersignalvec\odot\fouriersignalvec)$
for $\fouriersignalvec\define
\spacelaplacianevecmat\transpose\signalvec$ the frequency transform of
$\signalvec$ and $\frequencyweightvec\define[
  \frequencyweightfun(\spacelaplacianevalvertexnot{1}),\ldots,
  \frequencyweightfun(\spacelaplacianevalvertexnot{\vertexnum})
]\transpose$.  One can separately weight each frequency component by
selecting \cmt{Freq. response}$\frequencyweightvec$, which can be
thought of as the ``frequency response'' of the regularizer. For
instance, one may promote low pass estimates by setting the first
entries of $\frequencyweightvec$ to low values and the rest to high
values. 

\myitem\cmt{time-var}Inspired by this view, one may seek kernels
$\extendedfullkernelmat$ for which
\begin{align}
\label{eq:doublyselectivedef}
\regfun(\extendedrkhsvec)= \extendedrkhsvec\transpose
\extendedfullkernelmat\pinv \extendedrkhsvec= \tr{\frequencyweightmat\transpose(\fourierextendedsignalmat\odot\fourierextendedsignalmat)
}
\end{align}
where $\frequencyweightmat$ and $\fourierextendedsignalmat$ are
$\vertexnum \times \timenum$ matrices to be specified later
respectively containing the frequency response of the regularizer and
the bidimensional transform of $\signalfun$. The
$(\frequencyvertexind,\frequencytimeind)$-th entry of these matrices
corresponds to the $\frequencyvertexind$-th spatial frequency and
$\frequencytimeind$-th temporal frequency.  \cmt{term}Kernels
satisfying the second equality in \eqref{eq:doublyselectivedef} will
be termed \emph{doubly (frequency) selective}.  \cmt{example}Such
kernels preserve the flexibility of their counterparts for
time-invariant functions. For instance, if $\extendedfullkernelmat$
promotes doubly low-pass estimates, then the top left entries of
$\frequencyweightmat$ are small whereas the rest are large.



\end{myitemize}


\cmt{Form of doubly selective kernels}
\begin{myitemize}
\myitem\cmt{2d transform}To determine the form of a doubly-selective
kernel, let $\extendedsignalmat
\define[\signalvec\timenot{1},\ldots,\signalvec\timenot{\timenum}]$
and recall that a linear bidimensional transform 
can be expressed as
$\fourierextendedsignalmat\define\spacelaplacianevecmat\transpose\extendedsignalmat\timelaplacianevecmat$,
where the $\vertexnum\times\vertexnum$ matrix $\spacelaplacianevecmat$
and the $\timenum\times\timenum$ matrix $\timelaplacianevecmat$ stand for
orthogonal transformations along space and time, respectively.
\myitem\cmt{Kernel form}On the other hand, vectorizing the rightmost
term of~\eqref{eq:doublyselectivedef}~yields
\begin{align}
\label{eq:regularizer2d}
\regfun(\extendedrkhsvec)=
\extendedrkhsvec\transpose \extendedfullkernelmat\pinv \extendedrkhsvec
= \fourierextendedsignalvec\transpose \diag{\frequencyweightvec}
 \fourierextendedsignalvec
\end{align}
where 
$\frequencyweightvec\define \vect\{\frequencyweightmat\}$ and 
\begin{align}
 \fourierextendedsignalvec&\define
 \vect\{
\fourierextendedsignalmat\}= \vect\{
\spacelaplacianevecmat\transpose\extendedsignalmat\timelaplacianevecmat\}
=(\timelaplacianevecmat \otimes
\spacelaplacianevecmat)\transpose
\extendedsignalvec.
\label{eq:fourierextendedsignalvec}
\end{align}
Any doubly-selective kernel, or equivalently any kernel satisfying the
second equality of \eqref{eq:regularizer2d}, is therefore of the form
\begin{align}
\label{eq:frequencyinterpretablekernel}
\extendedfullkernelmat\pinv = 
(\timelaplacianevecmat \otimes
\spacelaplacianevecmat)
 \diag{\frequencyweightvec}
(\timelaplacianevecmat \otimes
\spacelaplacianevecmat)\transpose
\end{align}
for some orthogonal $\vertexnum\times\vertexnum$ matrix
$\timelaplacianevecmat$, some orthogonal $\timenum\times\timenum$ matrix $\spacelaplacianevecmat$,
and some entrywise non-negative vector~$\frequencyweightvec$. 
\end{myitemize}

\cmt{Laplacian doubly selec. kernels}Expression
\eqref{eq:frequencyinterpretablekernel} provides the general form of a
doubly-selective kernel, but a specific construction for
$\timelaplacianevecmat$, $\spacelaplacianevecmat$, and
$\frequencyweightvec$ capturing the spatiotemporal dynamics of
$\signalfun$ is still required. 
\begin{myitemize}
\myitem\cmt{roadmap}The next procedure serves  this purpose by
paralleling  the approach in Sec.~\ref{sec:background}. This involves
the following two steps.
\begin{myitemize}
\myitem\cmt{graph extension}\textbf{S1}: Since a Laplacian kernel
matrix shares eigenvectors with the Laplacian matrix, one should
construct an extended graph $\extendedgraph$ so that \change{its
  Laplacian matrix} $\extendedlaplacianmat$ is diagonalizable by
\change{a matrix of the form} $\timelaplacianevecmat \otimes
\spacelaplacianevecmat$ for some orthogonal
$\timelaplacianevecmat\in\rfield^{\timenum\times\timenum}$ and
$\spacelaplacianevecmat\in\rfield^{\vertexnum\times\vertexnum}$.
\myitem\cmt{spectral map}\textbf{S2}: One must design a spectral
weight map $\frequencyweightfun$ to obtain the eigenvalues of
$\extendedfullkernelmat$ from those of $\extendedlaplacianmat$.
\end{myitemize}
\end{myitemize}

\cmt{Kronecker kernels}
\begin{myitemize}
\myitem\cmt{S1: Extended graph}Regarding S1, \change{an explicit
  construction of an extended graph whose Laplacian matrix is
  diagonalizable by a matrix of the form $\timelaplacianevecmat
  \otimes \spacelaplacianevecmat$ with orthogonal
  $\timelaplacianevecmat\in\rfield^{\timenum\times\timenum}$ and
  $\spacelaplacianevecmat\in\rfield^{\vertexnum\times\vertexnum}$ is
  provided next. To this end,} consider the extended adjacency matrix
\begin{myitemize}
\myitem\cmt{Adjacency}
\begin{align}
\label{eq:kroneckeradjacency}
\extendedadjacencymat = \timeadjacencymat \oplus
\spaceadjacencymat 
\end{align}
where $\spaceadjacencymat $ is the given adjacency matrix of $\graph$
and the ${\timenum\times\timenum}$ adjacency matrix $
\timeadjacencymat$ is selected to capture temporal
dynamics. Specifically, with $\extendedadjacencymat$ as in
\eqref{eq:kroneckeradjacency}, the definition of extended adjacency
matrix in Sec.~\ref{sec:reconstruction} dictates that
\begin{myitemize}
\myitem the weight of the edge $(v\timevertexnot{\timeind}{n_1},
v\timevertexnot{\timeind}{n_2})$ for all $\timeind$ is given by the
$(n_1,n_2)$-th entry of $\spaceadjacencymat$, whereas \myitem the
weight of the edge $(v\timevertexnot{t_1}{\vertexind},
v\timevertexnot{t_2}{\vertexind})$ for all $\vertexind$ is given by
the $(t_1,t_2)$-th entry of $\timeadjacencymat$. A simple choice for
$\timeadjacencymat$ will be described later. 
\end{myitemize}
\myitem\cmt{Kronecker graphs}Note that~\eqref{eq:kroneckeradjacency}
differs from \emph{Kronecker graphs}~\cite{weichsel1962kronecker}, for
which $\extendedadjacencymat = \timeadjacencymat \otimes
\spaceadjacencymat$, although it can be interpreted as the
\emph{Cartesian graph} of $\vertexset$ and
$\{1,\ldots,\timenum\}$~\change{\cite{kashima2009pairwise,sandryhaila2014bigdata}}.
\change{Cartesian graphs have been considered in the graph signal
  processing literature for graph filtering and Fourier transforms of
  time-varying functions~\cite{sandryhaila2014bigdata}, but not for
   signal reconstruction.  }

\myitem\cmt{Laplacian}With $\extendedadjacencymat$ as in
\eqref{eq:kroneckeradjacency}, it can be readily seen that $
\extendedlaplacianmat \define
\diagnb\{\extendedadjacencymat\bm1\}-\extendedadjacencymat =
\timelaplacianmat \oplus \spacelaplacianmat $, where
$\timelaplacianmat \define
\diagnb\{\timeadjacencymat\bm1\}-\timeadjacencymat$ and
$\spacelaplacianmat \define
\diagnb\{\spaceadjacencymat\bm1\}-\spaceadjacencymat$ are the
Laplacian matrices associated with $\timeadjacencymat$ and
$\spaceadjacencymat$, respectively. 
\myitem\cmt{EVD}If $\timelaplacianmat = \timelaplacianevecmat
\diag{\timelaplacianevalvec} \timelaplacianevecmat\transpose$ and
$\spacelaplacianmat = \spacelaplacianevecmat
\diag{\spacelaplacianevalvec} \spacelaplacianevecmat\transpose$, then
\begin{align*}
\extendedlaplacianmat &= (\timelaplacianevecmat \otimes
\spacelaplacianevecmat)\left[
\diag{
\timelaplacianevalvec} \oplus\diag{
\spacelaplacianevalvec
}\right]
(\timelaplacianevecmat \otimes
\spacelaplacianevecmat)\transpose\\
&= (\timelaplacianevecmat \otimes
\spacelaplacianevecmat)
\diag{
\timelaplacianevalvec\otimes \bm 1_\vertexnum
+ \bm 1_\timenum \otimes
\spacelaplacianevalvec
}
(\timelaplacianevecmat \otimes
\spacelaplacianevecmat)\transpose.
\end{align*}
This expression reveals that the graph extension proposed in
\eqref{eq:kroneckeradjacency} indeed satisfies the objective of S1,
which requires the eigenvector matrix of $\extendedlaplacianmat$ to be
of the form $\timelaplacianevecmat \otimes \spacelaplacianevecmat$.
\change{Thus, it is always possible to construct a graph extension
  satisfying the goal of S1.  }

\end{myitemize}
\myitem\cmt{S2: Spectral weight map} 
\begin{myitemize}
\myitem\cmt{general form}For S2, one must construct a 
spectral map $\frequencyweightfun$ that yields
$\frequencyweightvec$ upon entrywise application to $
\timelaplacianevalvec\otimes \bm 1_\vertexnum + \bm 1_\timenum \otimes
\spacelaplacianevalvec $. To separately control the frequency response
along the spatial and temporal frequencies $\spacelaplacianeval$ and
$\timelaplacianeval$, such a map must take two
arguments as
$\frequencyweightfun(\timelaplacianeval,\spacelaplacianeval)$. This
results in $\frequencyweightvec
=\frequencyweightfun(\timelaplacianevalvec\otimes \bm 1_\vertexnum ,
\bm 1_\timenum \otimes \spacelaplacianevalvec)$ and \eqref{eq:frequencyinterpretablekernel} becomes
\begin{align}
\label{eq:frequencyinterpretablekernelform}
\extendedfullkernelmat\pinv =& 
(\timelaplacianevecmat \otimes
\spacelaplacianevecmat)
\\& \times \diag{\frequencyweightfun(\timelaplacianevalvec\otimes \bm 1_\vertexnum ,
\bm 1_\timenum \otimes \spacelaplacianevalvec)}
(\timelaplacianevecmat \otimes
\spacelaplacianevecmat)\transpose.
\nonumber
\end{align}
\myitem\cmt{term}Kernels of this form will be referred to as \emph{Kronecker
space-time} kernels. 
 \myitem\cmt{Possible choices}The transformation
$\frequencyweightfun$ can be selected in several ways. For instance,
\begin{myitemize}
\myitem\cmt{straightforward}the immediate construction at the
beginning of Sec.~\ref{sec:kernels} is recovered for
$\frequencyweightfun(\timelaplacianeval,\spacelaplacianeval)
=\frequencyweightfun(\timelaplacianeval+\spacelaplacianeval)$, with
$\frequencyweightfun(\laplacianeval)$ a one-dimensional spectral
weight map such as the ones in Table~\ref{tab:spectralweightfuns}.
\myitem\cmt{factorized $r$}Another possibility is to focus on
separable maps of the form
$\frequencyweightfun(\timelaplacianeval,\spacelaplacianeval)=
\timefrequencyweightfun(\timelaplacianeval)
\spacefrequencyweightfun(\spacelaplacianeval)$ where
$\timefrequencyweightfun$ and $\spacefrequencyweightfun$ denote
one-dimensional spectral maps. The resulting Kronecker kernel can
expressed as\footnote{The notion of Kronecker kernels together
  with~\eqref{eq:kernellaplacianprod} shows up in the literature of
  pairwise classification~\cite{kashima2009pairwise}, but the
  resemblance is merely illusional since the underlying kernel is a
  function of two \emph{pairs} of vertices.  }
\begin{align}
\label{eq:kernellaplacianprod}
\extendedfullkernelmat = \timefullkernelmat\otimes
\spacefullkernelmat
\end{align}
where $\timefullkernelmat
\define\timelaplacianevecmat
\diagnb\{\timefrequencyweightfun(\timelaplacianevalvec)\}
\timelaplacianevecmat\transpose$ and $\spacefullkernelmat
\define\spacelaplacianevecmat
\diagnb\{\spacefrequencyweightfun(\spacelaplacianevalvec)\}
\spacelaplacianevecmat\transpose$. For example, doubly bandlimited
estimates can be obtained by setting both $\timefullkernelmat$
and $\spacefullkernelmat$ to be bandlimited kernels
(Table~\ref{tab:spectralweightfuns}). 
\myitem\cmt{sum $r$}A further possibility  is to consider 
maps of the form $\frequencyweightfun(\timelaplacianeval,\spacelaplacianeval)=
\timefrequencyweightfun(\timelaplacianeval)+
\spacefrequencyweightfun(\spacelaplacianeval)$, which clearly result
in kernels of the form
\begin{align}
\label{eq:kernellaplaciansum}
\extendedfullkernelmat\pinv =
\timefullkernelmat\pinv\oplus \spacefullkernelmat\pinv.
\end{align}

\end{myitemize}

\end{myitemize}

\myitem\cmt{Link to online}To sum up, the proposed Kronecker kernels
\begin{myitemize}
\myitem arise from an intuitive graph extension and
\myitem can afford flexible adjustment of their frequency response.
\end{myitemize}Unfortunately, not any Kronecker
kernel is suitable for the online algorithm in
Sec.~\ref{sec:reconstruction} since the latter requires the inverse of
the kernel matrix $\extendedfullkernelmat$ to be block
tridiagonal. The rest of this section describes a subfamily of
Kronecker kernels that is suitable for this algorithm.

\myitem\cmt{Kronecker-online}
\begin{myitemize}
\myitem\cmt{construction}
Clearly, in order for
$\extendedfullkernelmat\pinv$ as in \eqref{eq:kernellaplacianprod} or
\eqref{eq:kernellaplaciansum} to be block tridiagonal, it is necessary
that $\timefullkernelmat\pinv$ be tridiagonal, i.e., the
$({\timeind,\timeindp})$-th entry of $\timefullkernelmat\pinv$ must be
zero if $|\timeind-\timeindp|> 1$.
\begin{myitemize}
\myitem\cmt{extended graph}Such a $\timefullkernelmat\pinv$ can be
obtained if, for instance, one sets the $(\timeind,\timeindp)$-th
entry of $\timeadjacencymat$ to be 0 unless
$|\timeind-\timeindp|=1$. In this extended graph construction, vertex
$v\timevertexnot{\timeind}{\vertexind}$, $1<\timeind<\timenum$, is
connected to $v\timevertexnot{\timeind-1}{\vertexind}$ and
$v\timevertexnot{\timeind+1}{\vertexind}$, which are its replicas in
adjacent time slots.  \myitem\cmt{spectral map}For
$\timefullkernelmat\pinv$ to be tridiagonal, one may set
$\timefrequencyweightfun(\timelaplacianeval)=\timelaplacianeval+\epsilon$,
where $\epsilon>0$ ensures that $\timefullkernelmat$ is invertible.

\end{myitemize}

\myitem\cmt{interpr.}Thus, the price to be paid for an online
implementation with the KKF from Sec.~\ref{sec:onlineestimator} is 
limited flexibility in specifying  the temporal frequency
response. \cmt{inherent KF}Note that this is not an intrinsic
limitation of the proposed algorithm, but it is inherent to the classical KF
as well; just recall that the latter assumes vector autoregressive
processes of order 1. \cmt{lowpass}In any case, the temporal frequency
response of a kernel for which
$(\timeadjacencymat)_{\timeind,\timeindp}=\delta[|\timeind-\timeindp|-1]$
can be obtained analytically by approximating the resulting Laplacian
$\timelaplacianmat$ for sufficiently large $\timenum$ with a circulant
matrix. This implies that (i) the eigenvectors in
$\timelaplacianevecmat$ are approximately those in the conventional
Fourier basis and therefore the notion of temporal frequency embodied in
$\timelaplacianevecmat$ preserves its conventional meaning; and (ii),
upon applying~\cite[Example 3]{romero2016multikernel}, the resulting
frequency response is low pass. Both (i) and (ii) are intuitively
reasonable. Thus, although the KKF solves only a subset of KRR
problems, this subset is of practical relevance.

\end{myitemize}
\begin{myremarkhere}
	\change{
		\myitem\cmt{matrix completion}
		\begin{myitemize}
\myitem\cmt{signals over space and time graph}In this paper, the rows of
$\extendedsignalmat$ can be thought of as graph functions over a graph
with adjacency matrix $\timeadjacencymat$, whereas the columns of
$\extendedsignalmat$ can be thought of as graph functions over the
graph with adjacency matrix $\spaceadjacencymat$.  \myitem\cmt{graph
  over that other dimension}In principle, each column of
$\extendedsignalmat$ does not need to correspond to a different time instant,
but  e.g. to a different movie in a  recommender
system application. The estimators~\eqref{eq:batchcriterion}-\eqref{eq:onlinecriterionsol}  can
therefore be used for matrix completion upon properly creating an extended
graph and graph kernel matrix.
  \myitem\cmt{space-space kernels}Towards this end, the space-time kernels
defined in
\eqref{eq:kernellaplacianprod} and \eqref{eq:kernellaplaciansum} 
readily generalize to space-space kernels that promote smoothness over
both graphs. 
		\end{myitemize}}
	\end{myremarkhere}

\end{myitemize}

\subsection{Space-time kernels for time-varying topologies}
\label{sec:tvtopologies}

\cmt{overview}For time-invariant topologies,
Sec.~\ref{sec:titopologies} proposed kernels that can be designed and
interpreted on a two-dimensional frequency plane. This section deals
with changing topologies, for which no bidimensional frequency notion
can be defined.

\cmt{spatial freq per $\timeind$}
\begin{myitemize}

\myitem\cmt{no 2D freq. notion} To recognize this claim, suppose that
$\spaceadjacencymat\timenot{\timeind}=\spaceadjacencymat$ remains
constant over $\timeind$ and recall that
$\spacelaplacianevec\vertexnot{\frequencyvertexind}$ is the
$\frequencyvertexind$-th eigenvector of $\spacelaplacianmat$ or,
equivalently, the $\frequencyvertexind$-th column of
$\spacelaplacianevecmat$. In this case, a bidimensional transform
exists and can be expressed as
$\fourierextendedsignalmat\define\spacelaplacianevecmat\transpose\extendedsignalmat\timelaplacianevecmat$,
whose $(\frequencyvertexind,\frequencytimeind)$-th entry corresponds
to the $\frequencyvertexind$-th spatial frequency and
$\frequencytimeind$-th temporal frequency. Fundamentally, the precise
meaning of the latter statement is that
$(\fourierextendedsignalmat)_{\frequencyvertexind,\frequencytimeind}$
is the $\frequencytimeind$-th temporal frequency component of the
$\frequencyvertexind$-th spatial frequency component of $\signalfun$,
i.e., the $\frequencytimeind$-th temporal frequency component of the
time series
$\{\fouriersignalfun\timevertexnot{\timeind}{\frequencyvertexind}\define(
\spacelaplacianevec\vertexnot{\frequencyvertexind})\transpose
\signalvec\timenot{\timeind}\}_{\timeind=1}^\timenum$, which is the
time evolution of the $\frequencyvertexind$-th \emph{spatial}
frequency component of $\signalfun$. However, for changing
topologies one cannot generally conceive the temporal evolution of a
specific spatial frequency component since the eigenvectors of
$\spacelaplacianmat\timenot{\timeind}$ generally differ from those of
$\spacelaplacianmat\timenot{\timeindp}$, thus precluding any natural
definition of the aforementioned sequence and therefore of a
bidimensional frequency transform.  \myitem\cmt{space freq. per
  $\timeind$}Nonetheless, it is shown next that the notion of spatial
frequency per slot $\timeind$ can still be utilized to design space-time
kernels for time-varying topologies.

\end{myitemize}

\cmt{Kernel design}
\begin{myitemize}
\myitem\cmt{extended graph construction}To this end, consider the
extended graph defined by~\eqref{eq:timevaryingextendedadjacencymat}
for arbitrary
$\timeconnectionmat\timenot{\timeind}\in\rfield_+^{\vertexnum\times\vertexnum}$. It
then follows that
\begin{align}
\nonumber
\extendedlaplacianmat
\define&
\diagnb\{\extendedadjacencymat \bm 1\}-\extendedadjacencymat
=
\bdiag{\spacelaplacianmat\timenot{1},\ldots,\spacelaplacianmat\timenot{\timenum}}
\\\nonumber&+\tridiag\big\{\diag{\timeconnectionmatcolsumvec\timenot{1}},\ldots,\diag{\timeconnectionmatcolsumvec\timenot{\timenum}};\\&-\timeconnectionmat\timenot{2},\ldots,-\timeconnectionmat\timenot{\timenum}\big\}
\label{eq:laplaciantimevarying}
\end{align}
where 
\begin{align*}
\timeconnectionmatcolsumvec\timenot{\timeind}
\define\begin{cases}
\timeconnectionmat\transpose\timenot{2}\bm
1&\text{if }\timeind=1\\
(\timeconnectionmat\transpose\timenot{\timeind+1}
+\timeconnectionmat\timenot{\timeind})
\bm
1&\text{if }1<\timeind<\timenum\\
\timeconnectionmat\timenot{\timenum}\bm
1&\text{if }\timeind=\timenum.
\end{cases}
\end{align*}
\myitem\cmt{Intuition}The rationale behind this graph extension is
that, for $\extendedlaplacianmat$ as in
\eqref{eq:laplaciantimevarying} and diagonal
$\{\timeconnectionmat\timenot{\timeind}\}_{\timeind=1}^\timenum$, one
can show  that
\begin{align}
\label{eq:regularizertimevaryinggraph}
\extendedsignalvec\transpose\extendedlaplacianmat\extendedsignalvec =&
\sum_{\timeind=1}^\timenum
\signalvec\transpose\timenot{\timeind}\spacelaplacianmat\timenot{\timeind}
\signalvec\timenot{\timeind}
\\&+
\sum_{\timeind=2}^\timenum(\signalvec\timenot{\timeind}-\signalvec\timenot{\timeind-1})\transpose
\timeconnectionmat\timenot{\timeind}(\signalvec\timenot{\timeind}-\signalvec\timenot{\timeind-1})\nonumber.
\end{align}
\begin{myitemize}
\myitem Clearly, the first and second sums on the right-hand side
respectively penalize spatial and temporal variations.  \myitem As a
special case, if one sets
$\timeconnectionmat\timenot{\timeind}=\timeconnection\bm I$
$\forall\timeind$ for some $\timeconnection>0$, the second  sum becomes
$\timeconnection\sum_{\timeind=2}^\timenum||\signalvec\timenot{\timeind}-\signalvec\timenot{\timeind-1}||_2^2$,
which promotes estimates with small changes over time. 
\end{myitemize}

\myitem\cmt{generalization} Applying the notion of Laplacian kernels
along the spatial dimension (see Sec.~\ref{sec:background}), but not
along time, suggests generalizing
\eqref{eq:regularizertimevaryinggraph} to obtain the regularizer
\begin{align}
\label{eq:regularizertimevaryingkernel}
\regfun(\extendedsignalvec)=&
\extendedsignalvec\transpose\extendedfullkernelmat\pinv\extendedsignalvec =
\sum_{\timeind=1}^\timenum
\signalvec\transpose\timenot{\timeind}\spacefullkernelmat\pinv\timenot{\timeind}
\signalvec\timenot{\timeind}
\\&+
\sum_{\timeind=2}^\timenum(\signalvec\timenot{\timeind}-\signalvec\timenot{\timeind-1})\transpose
\timeconnectionmat\timenot{\timeind}(\signalvec\timenot{\timeind}-\signalvec\timenot{\timeind-1})\nonumber
\end{align}
where $\spacefullkernelmat\pinv\timenot{\timeind}
=\frequencyweightfun_\timeind(\spacelaplacianmat\timenot{\timeind}) $,
$\timeind =1,\ldots,\timenum$ for
$\{\frequencyweightfun_\timeind\}_{\timeind=1}^\timenum$ a collection
of user-selected spectral maps such as those in
Table~\ref{tab:spectralweightfuns}. In that case, 
\eqref{eq:regularizertimevaryingkernel} corresponds to the kernel matrix
\begin{align}
\label{eq:kerneltimevarying}
\extendedfullkernelmat\pinv
=&\nonumber
\bdiag{\spacefullkernelmat\pinv\timenot{1},\ldots,\spacefullkernelmat\pinv\timenot{\timenum}}
\\&\nonumber
+\tridiag\big\{\diag{\timeconnectionmatcolsumvec\timenot{1}},\ldots,\diag{\timeconnectionmatcolsumvec\timenot{\timenum}};\\&-\timeconnectionmat\timenot{2},\ldots,-\timeconnectionmat\timenot{\timenum}\big\}.
\end{align}

\end{myitemize}

\begin{myitemize}
\myitem\cmt{critic}Although kernels of this form do not offer a
frequency-domain control of reconstruction along  time,
they still enjoy the  spatial flexibility of the
kernels in Sec.~\ref{sec:titopologies}.

\myitem\cmt{online}
\begin{myremarkhere}To guarantee that
$\extendedfullkernelmat\pinv$ in \eqref{eq:kerneltimevarying}
qualifies for online implementation, it suffices to guarantee that
$\extendedfullkernelmat$ is invertible since it is already block
tridiagonal. This holds e.g. if
$\spacefullkernelmat\timenot{\timeind}$ is invertible for all
$\timeind$.
\end{myremarkhere}

\end{myitemize}






\section{Simulated tests}
\label{sec:simulatedtests}


\newcommand{\dlsrstepsize}{\hc{\mu_\text{DLSR}}}
\newcommand{\lmsstepsize}{\hc{\mu_\text{LMS}}}
\newcommand{\dlsrbeta}{\hc{\beta_\text{DLSR}}}
\newcommand{\bandwidth}{\hc{B}}

\cmt{Sec. overview}This section compares the performance of the
proposed schemes with state-of-the-art alternatives and illustrates
some of the trade-offs inherent to time-varying function
reconstruction through real-data experiments.  \cmt{compared
  est.}Unless otherwise stated, the compared estimators include
\begin{myitemize}
\myitem distributed least squares reconstruction
(DLSR)~\cite{wang2015distributed} with step size $\dlsrstepsize$ and
parameter $\dlsrbeta$; \myitem the least mean-square (LMS) algorithm
in~\cite{lorenzo2016lms} with step size $\lmsstepsize$; \myitem
bandlimited instantaneous estimator (BL-IE), which results from
applying~\cite{narang2013localized,tsitsvero2016uncertainty,anis2016proxies}
separately per $\timeind$; \myitem KRR instantaneous estimator
(KRR-IE) reconstruction in \eqref{eq:timeagnostic} with a diffusion
kernel with parameter $\sigma$; \myitem and the proposed KKF
(Algorithms~\ref{algo:rec} and~\ref{algo:kalmanfilter}) with kernel
given by \eqref{eq:kerneltimevarying} for
$\timeconnectionmat\timenot{\timenum}=\timeconnection\bm I$ and
$\spacefullkernelmat\timenot{\timeind}$ a diffusion kernel with
parameter $\sigma$.
\end{myitemize}
DLSR, LMS, and BL-IE also use a bandwidth
parameter~$\bandwidth$.

\cmt{Data set 1: Temperature evolution across contiguous USA} The
first data set comprises hourly temperature measurements at
$\vertexnum=109$ stations across the continental U.S. in
2010~\cite{USATemp}. Temperature reconstruction has been extensively
employed in the literature to analyze the performance of inference
tools over graphs (see
e.g.~\cite{mei2016causal,wang2015distributed,bach2004learning}).
\begin{myitemize}\myitem\cmt{Configuration}\begin{myitemize}
    \myitem\cmt{adjacency}A time-invariant graph was constructed
    following the approach in~\cite{mei2016causal} with 7 nearest
    neighbors, which relies on geographical
    distances.
    \myitem\cmt{signal} Function
    $\signalfun\timevertexnot{\timeind}{\vertexind}$ represents the
    temperature at the $\vertexind$-th  station and
    $\timeind$-th sampling instant. In the first experiment, the
    latter corresponds to the $\timeind$-th hour, whereas for the
    rest, it corresponds to the temperature at 12:00 PM of the
    $\timeind$-th day.
       
  \end{myitemize} 

      \myitem\cmt{Experiment 1:Tracking Hourly Temperature 
	\ra Fig.~\ref{fig:realTempTrack}}  

	\begin{figure}
	    \centering
	    \includegraphics[width=8.5cm]{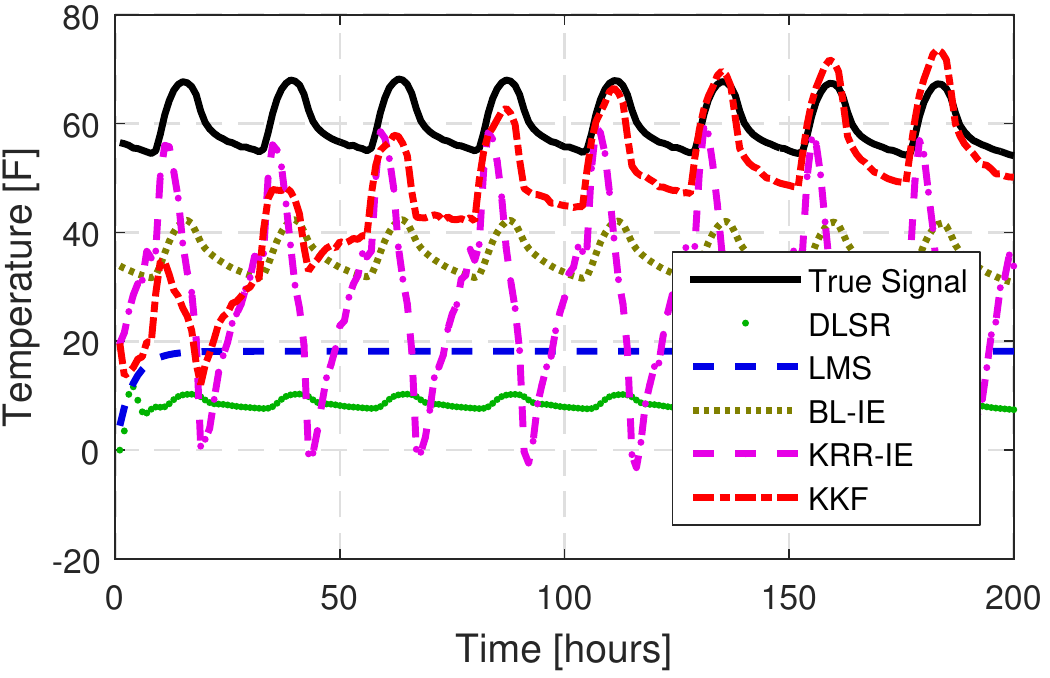}
	  \caption{True temperature and estimates across time at a
            randomly picked unobserved station ($\regpar=10^{-7}$,
            $\sigma=1.8$, $\timeconnection =0.01$, $\dlsrstepsize
            =1.2$, $\dlsrbeta=0.5$, $\lmsstepsize =0.6$, $B=2$).}
	    \label{fig:realTempTrack}
	\end{figure}

      \begin{myitemize}
	\myitem\cmt{goal}Fig.~\ref{fig:realTempTrack} depicts the true
        temperature measured at an unobserved randomly picked station
        over the first 200 hours of 2010 along with its estimates for
        a typical realization of the time-invariant sampling set
        $\sampleset=\sampleset\timenot{\timeind},~\forall\timeind,$
        drawn at random within all sampling sets with $\samplenum=44$
        elements.  \myitem\cmt{Observations}Different from
        instantaneous  alternatives, whose error does not
        decrease with time, KKF is observed to successfully leverage
        time dynamics to track the temperature at the unobserved
        station.  On the other hand, DLSR and LMS are unable to track
        the rapid variations of $\signalfun$ since their design
        assumes slowly changing functions.
      \end{myitemize}

\cmt{NMSE def}The next experiments compare the cumulative normalized
mean-square error (NMSE), defined as
  \begin{align*}
    \text{NMSE}(\timeind,\{\sampleset\timenot{\tau}\}_{\tau=1}^\timeind):=\frac{\sum_{\tau=1}^{\timeind}
      ||\samplemat^c\timenot{\tau}(\signalvec\timenot{\tau}
      -\signalestvec\timenot{\tau|\tau})||^2_2}
	 {\sum_{\tau=1}^{\timeind}	 
	   ||\samplemat^c\timenot{\tau}\signalvec
	   \timenot{\tau}||^2_2}
  \end{align*}
  where $\samplemat^c\timenot{\tau}$ is an
  $\vertexnum-\samplenum\timenot{\tau}\times \vertexnum$ matrix
  comprising the rows of $\identitymat_\vertexnum$ whose indices are
  not in $\sampleset\timenot{\timeind}$.

      \myitem\cmt{Experiment 2: error vs. time 
	Fig.~\ref{fig:realTemp} }
      \begin{myitemize}
	\begin{figure}
	    \centering
	    \includegraphics[width=8.5cm]{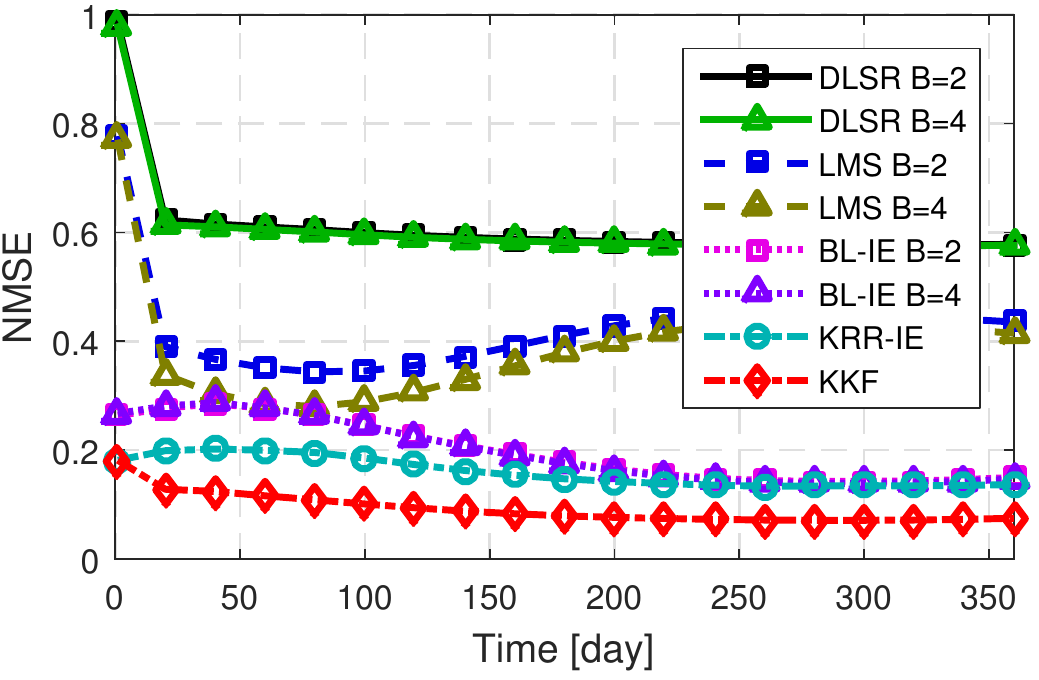}
	    
	   
	  
	  \caption{NMSE of daily temperature estimates over 2010.
	    ($\regpar=10^{-7}$,
	    $\sigma=1.8$,
	    $\timeconnection =0.01$,
	    $\dlsrstepsize =1.2$,
	    $\dlsrbeta=0.5$,
	    $\lmsstepsize =0.6$).
}
	 \label{fig:realTemp}
	\end{figure}

	\myitem\cmt{Goal}Fig.~\ref{fig:realTemp} shows the NMSE for
        $\sampleset\timenot{\timeind} = \sampleset,~\forall\timeind$,
        averaged over all possible $\sampleset$ with $\samplenum=44$
        elements.  \myitem\cmt{Observations}It is observed that the
        instantaneous estimators outperform DLSR and LMS, which can
        only cope with slow variations of $\signalfun$. Furthermore,
        the error of KKF is half the error of the nearest alternative,
        demonstrating the importance of exploiting time dynamics.
      \end{myitemize}
      
      \myitem\cmt{Experiment 3:  error vs sampling 
	size\ra Fig.~\ref{fig:realTempIncrSamp}}  
      \begin{myitemize}
	\begin{figure}
	    \centering
	    \includegraphics[width=8.5cm]{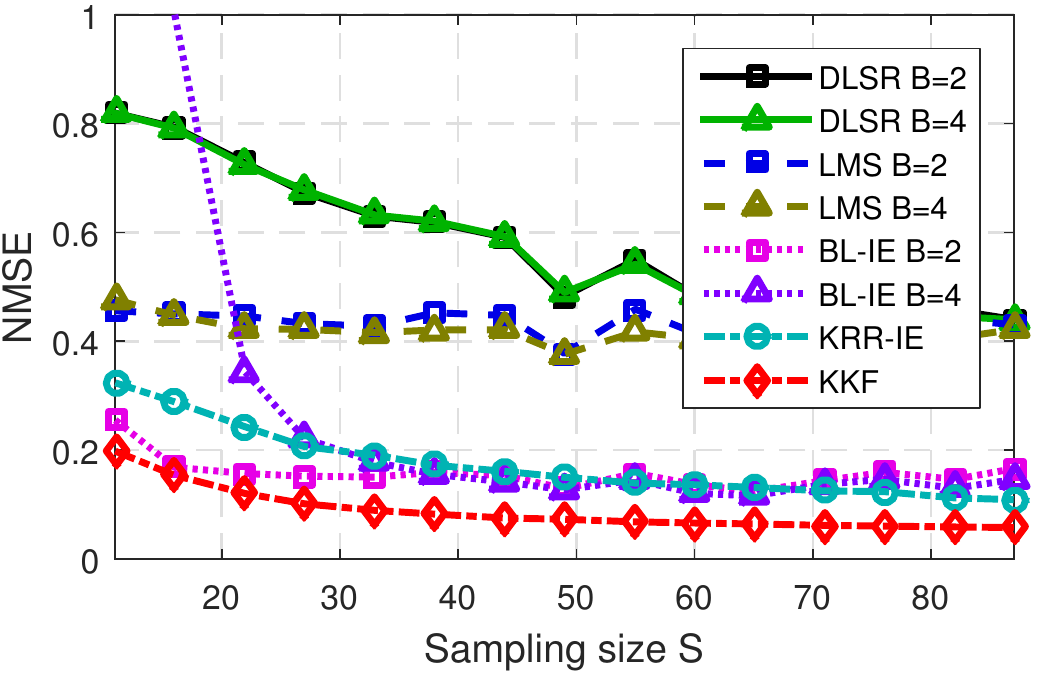}
	    
	  
	  \caption{NMSE for increasing sampling size
	  	($\regpar=10^{-7}$,
	  	$\sigma=1.6$,
	  	$\timeconnection =0.01$,
	  	$\dlsrstepsize =1.2$,
	  	$\dlsrbeta=0.5$,
	  	$\lmsstepsize =0.6$).}
	    \label{fig:realTempIncrSamp}
	\end{figure}

	\myitem\cmt{Goal}Fig.~\ref{fig:realTempIncrSamp} shows
        the impact of the number of observed vertices $\samplenum$ in
        $\text{NMSE}(\timenum,\{\sampleset\timenot{\timeindaux}\}_{\tau=1}^\timeind)$,
        with $\timenum=365$ days, averaged over all sets
        $\sampleset\timenot{\timeindaux}=\sampleset~\forall\timeindaux$
        with $\samplenum$ elements.  \myitem\cmt{Observations}Observe
        that KKF consistently outperforms all alternatives. Still, the
        advantage of KKF over KRR-IE is more pronounced for small
        $\samplenum$, since in that case exploiting the time dynamics
        is more critical.
      \end{myitemize}

      \myitem\cmt{Experiment 4: error vs offdiag scaling \ra
        Fig.~\ref{fig:paramselection}}
      \begin{myitemize}
	\begin{figure}
	    \centering
	    \includegraphics[width=8.5cm]{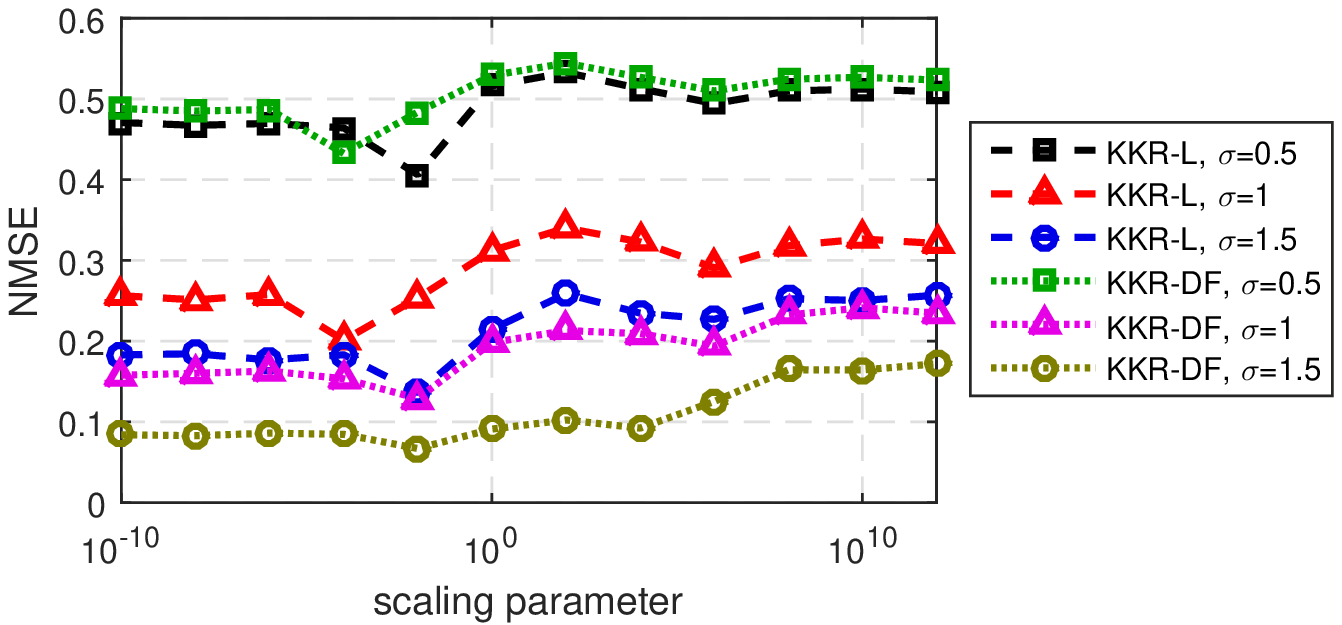}
	  \caption{NMSE for different kernels vs. scale parameter $
            \timeconnection$ ($\regpar=10^{-7}$).
            }
	    \label{fig:paramselection}
	\end{figure}
	\myitem\cmt{Goal}To illustrate the trade-off between
        reliance on temporal versus spatial information, the next
        experiment analyzes the effects of the scaling parameter $
        \timeconnection$ in the kernel adopted by KRR
        (cf.~\eqref{eq:kerneltimevarying}). A large value of
        $\timeconnection$ leads to an estimator that relies more
        heavily on time dynamics and vice versa.
        Fig.~\ref{fig:paramselection} shows
        $\text{NMSE}(\timenum,\{\sampleset\timenot{\tau}\}_{\tau=1}^\timeind)$,
        with $\timenum=100$ days, averaged \change{over all sets
        $\sampleset\timenot\tau=\sampleset~\forall\tau$ with $\samplenum=44$
        elements.}  \myitem\cmt{Kernels used}The kernel
        in~\eqref{eq:kerneltimevarying} is adopted with
        $\spacefullkernelmat\timenot{\timeind}$ being the regularized
        Laplacian (KKF-L) or diffusion kernels (KKF-DF) from Table
        \ref{tab:spectralweightfuns}, while
        $\timeconnectionmat\timenot{\timeind}=\timeconnection\bm I$.
        \myitem\cmt{Result}It is observed that there exists an optimum
        value for $\timeconnection$ which leads to the best
        reconstruction performance. This corresponds to the optimal
        trade-off point between reliance on temporal and spatial
        information. \change{The optimal NMSE is achieved by a diffusion kernel with $\sigma=1.5$ and 
        $\timeconnection=0.01$.}


      \end{myitemize}

\end{myitemize}

\cmt{Data set 2: Economic sectors} The second data set is provided by
the Bureau of Economic Analysis of the U.S.  Department of Commerce
and contains the annual investments between each pair of sectors among
\change{$\vertexnum=61$ economic sectors in the interval 1997-2014~\cite{BEconanalysis}.}
\begin{myitemize}\myitem\cmt{Configuration}\begin{myitemize}
		\myitem\cmt{adjacency}\change{Each entry 
		$\spaceadjacencymatentry\timevertexvertexnot{\timeind}{\vertexind}{\vertexindp}$ of 
		$\spaceadjacencymat\timenot{\timeind}$ contains 
     the 
     investment in trillions of dollars between sectors $\vertexind$ and $\vertexindp$ for the year $1995 
     +2\timeind$ 
    with  $\timeind=1,2, \ldots, \timenum$, where $\timenum=9$.  
     DLSR and LMS adopt
    $\spaceadjacencymat=({1}/{\timenum})\sum_{\tau=1}^{\timenum}
    \spaceadjacencymat\timenot{\tau}$ since they cannot handle
    time-varying topologies.  \myitem\cmt{signal} The value
    $\signalfun\timevertexnot{\timeind}{\vertexind}$ corresponds to
    the total production of the $\vertexind$-th sector in year $1996 +2\timeind,~~\timeind=1,2, \ldots, 
    \timenum$. The sampling interval was set to two years, so that  disjoint subsets of years are used  for 
    generating the signal and constructing the graphs.}

      \end{myitemize}

      \myitem\cmt{Experiment 1: error
        vs. time~\ref{fig:econnmseacrosstime} }  
      \begin{myitemize}
	\begin{figure}
	    \centering
	    \includegraphics[width=8.5cm]{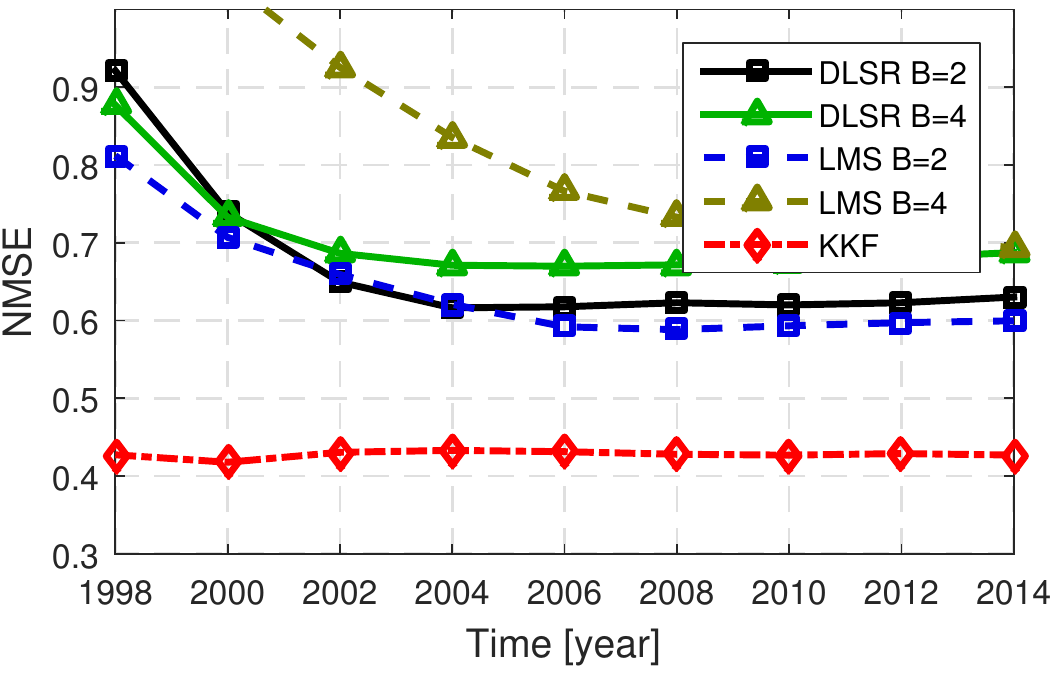}
	  \caption{NMSE for the economic sectors data set
            ($\sigma=5.2$, $\regpar=10^{-4}$, $\timeconnection =0.01$, $\dlsrstepsize=1.2$,
            $\dlsrbeta =0.5$, $\lmsstepsize =0.6$).}
	    \label{fig:econnmseacrosstime}
	\end{figure}
			
		

	    
	  
	
	\myitem\cmt{Goal}The next experiment demonstrates the ability
        of KKF to handle time-varying topologies.
        \myitem\cmt{Y-axis}To this end,
        Fig.~\ref{fig:econnmseacrosstime} plots
        $\text{NMSE}(\timeind,\{\sampleset\timenot{\tau}\}_{\tau=1}^\timeind)$,
        averaged over all sets
        $\sampleset\timenot\timeind=\sampleset,~\forall\timeind,$ with
        $\samplenum=37$ elements.  \myitem\cmt{Kernels}KKF utilizes
        the kernel in~\eqref{eq:kerneltimevarying} with
        $\spacefullkernelmat\timenot\timeind$ a diffusion kernel
        constructed from $\spacelaplacianmat\timenot\timeind$ per
        $\timeind$ and  $\timeconnectionmat\timenot{\timeind}=
        \timeconnection\bm I,~\forall\timeind$.
        \myitem\cmt{Observations}Again,
        Fig.~\ref{fig:econnmseacrosstime} showcases the superior
        performance of the proposed KKF, whose error is significantly less than the error
        of competing alternatives.

      \end{myitemize}
\change{      
\cmt{Data set 3: Brain signals}The third data set  is obtained from  an epilepsy 
study~\cite{kramer2008seizure}.  
Diagnosis of epilepsy is heavily based on analysis of ECoG data; see Sec.~\ref{sec:intro}.
}
\begin{myitemize}
\change{
	\myitem\cmt{Configuration}The next experiments utilize the ECoG time series 
	 obtained in~\cite{kramer2008seizure} from 
	$\vertexnum=76$ 
	electrodes implanted in a patient's brain before and after the onset of a 
	seizure.
}
\change{
	\begin{myitemize}
		\myitem\cmt{adjacency}A symmetric time-invariant
                adjacency matrix $\spaceadjacencymat$ was obtained
                using the method in~\cite{shen2016nonlinear} with ECoG
                data before the onset of the seizure.
                \myitem\cmt{signal}Function
                $\signalfun\timevertexnot{\timeind}{\vertexind}$
                comprises the electrical signal at the $\vertexind$-th
                electrode and $\timeind$-th sampling instant after the
                onset of the seizure, for a period of $\timenum=250$
                samples. The values of
                $\signalfun\timevertexnot{\timeind}{\vertexind}$ were
                normalized by subtracting the temporal mean of each
                time series  before the onset of the
                seizure.
	\end{myitemize}
	\myitem\cmt{Experiment 1: error vs. time}
}
	\begin{myitemize}
		\begin{figure}
			\centering
			\includegraphics[width=8.5cm]{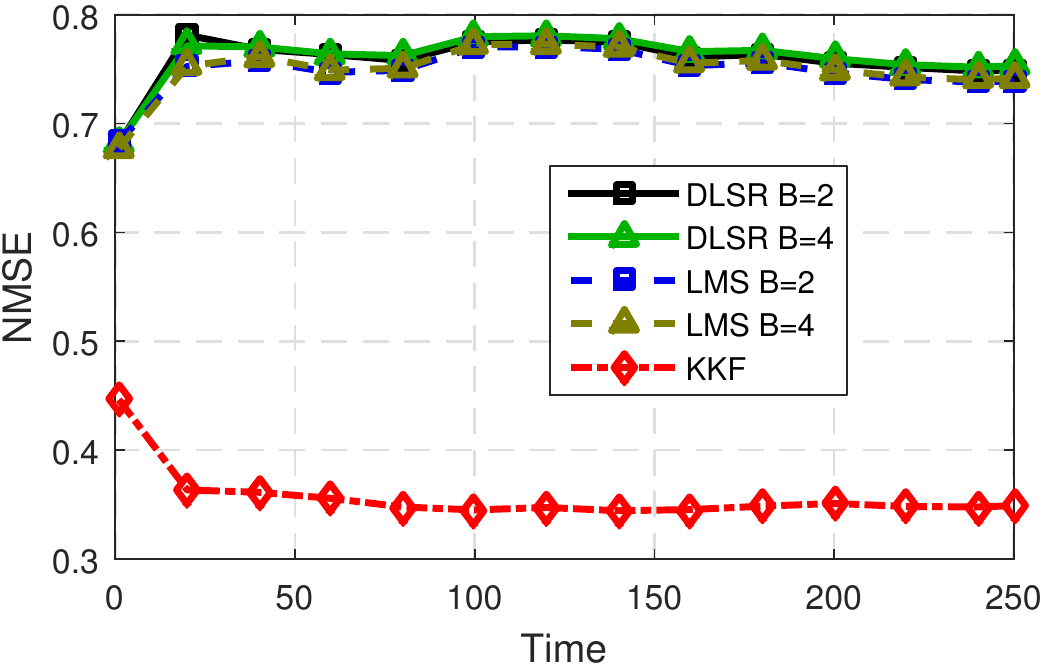}
			\caption{NMSE for the ECoG data set
				($\sigma=1.2$, $\regpar=10^{-4}$, $\dlsrstepsize=1.2$, $\timeconnection =0.01$,
				$\dlsrbeta =0.5$, $\lmsstepsize =0.6$).}
			\label{fig:brainsignalacrosstime}
		\end{figure}
\change{
		\myitem\cmt{Goal}The goal of the experiment is to illustrate the 
		reconstruction performance of the proposed KKF in capturing the complex 
		spatio-temporal dynamics of brain signals.
}

\change{		\myitem\cmt{Plot}Fig.~\ref{fig:brainsignalacrosstime} depicts the 
		$\text{NMSE}(\timeind,\{\sampleset\timenot{\tau}\}_{\tau=1}^\timeind)$,
		averaged over all sets
		$\sampleset\timenot\timeind=\sampleset,~\forall\timeind,$ of size
		$\samplenum=53$.  
		\myitem\cmt{Kernel}For the proposed KKF, a space-time kernel was 
		created using \eqref{eq:kerneltimevarying} with a time-invariant 
		diffusion kernel
		$\spacefullkernelmat$  generated from 
		$\spacelaplacianmat$, and a time-invariant
		$\timeconnectionmat=
		\timeconnection\bm I$.
		\myitem\cmt{Result}Fig.~\ref{fig:brainsignalacrosstime}
                showcases the superior  reconstruction performance of
                the KKF among 
	    competing approaches, even with 
		a small number of samples. This result suggests that the ECoG 
		diagnosis technique could be efficiently conducted even with a smaller 
		number of intracranial electrodes, which may have a great impact on the 
		patient's experience.		
}
	\end{myitemize}	
\end{myitemize}

\end{myitemize}

\section{Conclusions}
\label{sec:conclusions}

This paper investigated kernel-based reconstruction of 
space-time functions on graphs. The adopted approach relied on the
construction of an extended graph, which regards the time dimension
just as a spatial dimension. Several kernel designs were introduced
together with batch and online function estimators. Future research
will deal with multi-kernel and distributed versions of the proposed
algorithms.

\appendix

\subsection{Proof of Lemma~\ref{lem:kf-krreq}}
\label{proof:lemma}
	\begin{myitemize} \myitem\cmt{lemma to relate with state space
	    space model}
\change{Start by noting that
			\begin{align}
			&\sum_{\tau=2}^{\timenum}||\rkhsvec\timenot{\tau} -
			\transitionmat\timenot{\tau}\rkhsvec\timenot{\tau-1}||^2_{
				\plantnoisekernelmat\timenot{\tau}}\nonumber\\
			=&\sum_{\tau=2}^{\timenum}(\rkhsvec\timenot{\tau}
			-
			\transitionmat\timenot{\tau}\rkhsvec\timenot{\tau-1})\transpose
			\plantnoisekernelmat\inv\timenot{\tau}(\rkhsvec\timenot{\tau}
			-
			\transitionmat\timenot{\tau}\rkhsvec\timenot{\tau-1})\nonumber\\
			=&\sum_{\tau=2}^{\timenum}\bigg(\rkhsvec\transpose\timenot{\tau}
			\plantnoisekernelmat\inv\timenot{\tau}\rkhsvec\timenot{\tau}
			-\rkhsvec\transpose\timenot{\tau}
			\plantnoisekernelmat\inv\timenot{\tau}\transitionmat\timenot{\tau}
			\rkhsvec\timenot{\tau-1}
			\nonumber\\
			&-\rkhsvec\transpose\timenot{\tau-1}
			\transitionmat\transpose\timenot{\tau} 
			\plantnoisekernelmat\inv\timenot{\tau}
			\rkhsvec\timenot{\tau}\nonumber\\					 
			&+\rkhsvec\transpose\timenot{\tau\!-\!1}
			\transitionmat\transpose\timenot{\tau}
			\plantnoisekernelmat\inv\timenot{\tau}				
			\transitionmat\timenot{\tau}\rkhsvec\timenot{\tau\!-\!1}\bigg)
			\nonumber\\
			=&\sum_{\tau=2}^{\timenum}\bigg(\rkhsvec\transpose\timenot{\tau-1}
			\big(\plantnoisekernelmat\inv\timenot{\tau-1}+
				\transitionmat\transpose\timenot{\tau}
				\plantnoisekernelmat\inv\timenot{\tau}				
				\transitionmat\timenot{\tau}\big)\rkhsvec\timenot{\tau-1}\nonumber\\
			&-\rkhsvec\transpose\timenot{\tau}
			\plantnoisekernelmat\inv\timenot{\tau}\transitionmat\timenot{\tau}
			\rkhsvec\timenot{\tau-1}\nonumber\\					 
			&
			-\rkhsvec\transpose\timenot{\tau-1}
			\transitionmat\transpose\timenot{\tau} 
			\plantnoisekernelmat\inv\timenot{\tau}
			\rkhsvec\timenot{\tau}					 \bigg)\nonumber\\
			&-\rkhsvec\transpose\timenot{1}\plantnoisekernelmat\inv\timenot{1}
			\rkhsvec\timenot{1}
			+\rkhsvec\transpose\timenot{\timenum}\plantnoisekernelmat\inv\timenot{\timenum}
			\rkhsvec\timenot{\timenum}.
			\nonumber
\end{align}
Recursively solving the equations in Algorithm \ref{algo:rec} for
$\{\transitionmat\timenot{\tau}\}_{\tau=2}^\timenum$
and~$\{\plantnoisekernelmat\timenot{\tau}\}_{\tau=1}^{\timenum}$
yields
\begin{align}
  \sum_{\tau=2}^{\timenum}&||\rkhsvec\timenot{\tau} -
			\transitionmat\timenot{\tau}\rkhsvec\timenot{\tau-1}||^2_{
				\plantnoisekernelmat\timenot{\tau}}\nonumber\\
			=&\sum_{\tau=2}^{\timenum}\bigg(\rkhsvec\transpose\timenot{\tau-1}
			\kernelinvondiagonalmat\timenot{\tau-1}\rkhsvec\timenot{\tau-1}\nonumber\\
			&+\rkhsvec\transpose\timenot{\tau}
		    \kernelinvoffdiagonalmat\timenot{\tau}
			\rkhsvec\timenot{\tau-1}
			+\rkhsvec\transpose\timenot{\tau-1}
			\kernelinvoffdiagonalmat\transpose\timenot{\tau}
			\rkhsvec\timenot{\tau}					 \bigg)\nonumber\\
			&-\rkhsvec\transpose\timenot{1}\plantnoisekernelmat\inv\timenot{1}
			\rkhsvec\timenot{1}
			+\rkhsvec\transpose\timenot{\timenum}\kernelinvondiagonalmat\timenot{\timenum}
			\rkhsvec\timenot{\timenum}
			\nonumber
			\\
			=&\extendedrkhsvec\transpose
			\extendedfullkernelmat\inv
			\extendedrkhsvec	
			-\rkhsvec\transpose\timenot{1}\plantnoisekernelmat\inv\timenot{1}
			\rkhsvec\timenot{1}
			\label{eq:kernelsub}
			\end{align}
where the
			 last equality follows 
			from~\eqref{eq:invblocktriagonal}. After substituting 
			\eqref{eq:kernelsub} into \eqref{eq:kfobj} and recognizing that the first summand in 
			\eqref{eq:kfobj} equals $||\extendedobservationvec\timenot{\timeind} -
			\extendedsamplemat\timenot{\timeind}\extendedrkhsvec||^2_{
				\samplenumdiagmat\timenot{\timeind}
			}$, expression~\eqref{eq:onlinecriterion} is recovered, which concludes the proof.
                        }
			%
        \end{myitemize}

        \subsection{Proof of Theorem~\ref{prop:KFanytridiag}}
        \label{proof:theorem}

        \begin{myitemize}
\change{ \myitem\cmt{establish equivelant problem of increasing
    size}The first step is to simplify the objective
  in~\eqref{eq:kfobj}. To this end, note that minimizing~\eqref{eq:kfobj} with
  respect to $\{\rkhsvec[\tau]\}_{\tau=\timeindp+1}^\timenum $ for any
  $\timeind$ and $\timeindp$ such that $\timeindp\ge\timeind$ yields
\begin{subequations}
			\begin{align}
\label{eq:predictiontau}
                        \signalestvec\timegiventimenot{\tau}{\timeind}=& 
				\transitionmat\timenot{\tau}
				\signalestvec\timegiventimenot{\tau-1}{\timeind},~\tau=\timeindp+1,\ldots,\timenum,\\
			\{\rkhsestvec[\tau|\timeind]\}_{\tau=1}^{\timeindp}
			=& \underset{\{\rkhsvec[\tau]\}_{\tau=1}^{\timeindp}}{\argmin}
			~\sum_{\tau=1}^{\timeind}\frac{1}{\observationnoisevar\timenot{\tau}}||\observationvec\timenot{\tau}
			-
			\samplemat\timenot{\tau}\rkhsvec\timenot{\tau} ||^2\nonumber\\
			&+\sum_{\tau=2}^{\timeindp}||\rkhsvec\timenot{\tau}-\nonumber
			~\transitionmat\timenot{\tau}\rkhsvec\timenot{\tau-1}||^2_ 
			{\plantnoisekernelmat\timenot{\tau}}\\
			&+\rkhsvec\transpose\timenot{1}\plantnoisekernelmat\inv\timenot{1}
			\rkhsvec\timenot{1}.
			\label{eq:truekfobj}		
			\end{align}
\end{subequations}
}

 \myitem\cmt{transform
				to book form}\change{			The goal is therefore to show that the
				$\timeind$-th iteration of
				Algorithm \ref{algo:kalmanfilter}
				returns
				$\signalestvec\timegiventimenot{\timeind}{\timeind}$
				as given by~\eqref{eq:truekfobj}.  To simplify notation,
				collect the function values up to time
				$\timeind$ as
				$\extendedgrowingsignalvec\timenot{\timeind} \define[\rkhsvec\transpose\timenot{1}, \rkhsvec\transpose\timenot{2},\ldots, \rkhsvec\transpose\timenot{\timeind}]\transpose \in\rfield^{\vertexnum\timeind}$
				and their estimates given observations
				up to time $\timeindp$ as
				$\extendedgrowingsignalestvec\timegiventimenot{\timeind}{\timeindp} \define[\rkhsestvec\transpose\timegiventimenot{1}{\timeindp}, \rkhsestvec\transpose\timegiventimenot{2}{\timeindp},\ldots, \rkhsestvec\transpose\timegiventimenot{\timeind}{\timeindp}]\transpose \in\rfield^{\vertexnum\timeind}$.
				The rest of the proof proceeds along
				the lines
				in~\cite[Ch. 17]{strang1997linear} by
				expressing
				$\extendedgrowingsignalestvec\timegiventimenot{\timeind}{\timeind}$
				and
				$\extendedgrowingsignalestvec\timegiventimenot{\timeind}{\timeind-1}$
				as the solutions to two least-squares
				problems.  To
				this end, define the
				$Nt+\extendedsamplenum\timenot{\timeind}\times
				Nt$
				matrix \small{\begin{align} \!\extendedobs\timenot{\timeind}\!\define\!  \left[\!\begin{array}{ccccccc} \!\identitymat_\vertexnum
				& \bm 0& \bm 0& \ldots & \bm 0& \bm
				0& \bm 0 \\ \!\samplemat\timenot{1}
				& \bm 0& \bm 0& \ldots & \bm 0& \bm
				0& \bm
				0 \\ \!-\transitionmat\timenot{2}
				& \identitymat_\vertexnum & \bm
				0& \ldots & \bm 0& \bm 0& \bm
				0\\ \!\bm 0
				& \samplemat\timenot{2}& \bm 0
				& \ldots& \bm 0& \bm 0& \bm
				0\\ \!\vdots & \vdots & \vdots
				& \ddots & \vdots & \vdots
				& \vdots\\ \bm 0 & \bm 0 & \bm
				0& \ldots & \bm
				0& \samplemat\timenot{\timeind-1} &\bm
				0 \\ \!\bm 0 & \bm 0 & \bm 0& \ldots
				& \bm 0&
				-\transitionmat\timenot{\timeind}
				&\identitymat_\vertexnum \\ \!\bm 0
				& \bm 0 & \bm 0& \ldots & \bm 0 &\bm
				0& \samplemat\timenot{\timeind} \\ \end{array} \right] \end{align}}\normalsize
				the
				${Nt+\extendedsamplenum\timenot{\timeind}\times
				Nt+\extendedsamplenum\timenot{\timeind}}$
				matrix
				$\extendedsigma\timenot{\timeind} \define\bdiagnb\{\plantnoisekernelmat\timenot{1}, \observationnoisevar\timenot{1} \identitymat_{\samplenum\timenot{1}}, \plantnoisekernelmat\timenot{2}, \observationnoisevar\timenot{2} \identitymat_{\samplenum\timenot{2}}, \ldots, \observationnoisevar\timenot{t-1} \identitymat_{\samplenum\timenot{t-1}},
				$
				$ \plantnoisekernelmat\timenot{t}, \observationnoisevar\timenot{t} \identitymat_{\samplenum\timenot{t}}\}$,
				and note from
}
		\begin{myitemize} \myitem\cmt{establish equivelant
			problem as book}\change{\eqref{eq:truekfobj} that \begin{align} \extendedgrowingsignalestvec\timegiventimenot{\timeind}{\timeind}= \underset{\extendedgrowingsignalvec\timenot{\timeind}
			}{\argmin}||\extendedgrowingobsvec\timenot{\timeind}
			- \extendedobs\timenot{\timeind}\extendedgrowingsignalvec\timenot{\timeind}
			||^2_{\extendedsigma\timenot{\timeind}} \label{eq:transtruekfobj} \end{align}
			where
			$\extendedgrowingobsvec\timenot{\timeind} \define[\bm
			0_\vertexnum\transpose,\observationvec\transpose\timenot{1}, \bm
			0_\vertexnum\transpose,\observationvec\transpose\timenot{2}, \bm
			0_\vertexnum\transpose,\ldots, \bm
			0_\vertexnum\transpose, \observationvec\transpose\timenot{\timeind}]\transpose \in\rfield^{\vertexnum\timeind+\extendedsamplenum\timenot{\timeind}}$.
			Indeed, expression~\eqref{eq:transtruekfobj}
			corresponds to the weighted least-squares
			solution to  \begin{align} \extendedgrowingobsvec\timenot{\timeind}= \extendedobs\timenot{\timeind}\extendedgrowingsignalvec\timenot{\timeind}+ \extendedgrowingnoisevec\timenot{\timeind} \label{eq:systemofeq} \end{align}
			where
			$\extendedgrowingnoisevec\timenot{\timeind}\in\rfield^{
			Nt+\extendedsamplenum\timenot{\timeind}}$ is
			an error vector, and admits the closed-form solution \begin{align} \extendedgrowingsignalestvec\timegiventimenot{\timeind}{\timeind}=(\extendedobs\transpose \timenot{\timeind}\extendedsigma\inv\timenot{\timeind}\extendedobs \timenot{\timeind})\inv\extendedobs\transpose \timenot{\timeind}\extendedsigma\inv\timenot{\timeind}\extendedgrowingobsvec \timenot{\timeind}.  \label{eq:closedformtot}\end{align}
			%
			%
			%
}

\change{
                        Similarly, define the
			 $Nt+\extendedsamplenum\timenot{\timeind-1}\times
			 Nt$ matrix \small{\begin{align}
			 &\extendedtrs\timenot{\timeind}\define\nonumber\\
			 &\left[ \begin{array}{ccccccc} \identitymat_\vertexnum
			 & \bm 0& \bm 0& \ldots & \bm 0& \bm 0& \bm
			 0 \\ \samplemat\timenot{1} & \bm 0& \bm
			 0& \ldots & \bm 0& \bm 0& \bm 0 \\
			 -\transitionmat\timenot{2}
			 & \identitymat_\vertexnum & \bm 0& \ldots
			 & \bm 0& \bm 0& \bm 0\\ \bm 0
			 & \samplemat\timenot{2}& \bm 0 & \ldots& \bm
			 0& \bm 0& \bm 0\\ \vdots & \vdots & \vdots
			 & \ddots & \vdots & \vdots & \vdots\\ \bm 0
			 & \bm 0 & \bm 0& \ldots &
			 -\transitionmat\timenot{\!\timeind-1}
			 &\identitymat_\vertexnum& \bm 0\\ \bm 0 & \bm
			 0 & \bm 0& \ldots & \bm
			 0& \samplemat\timenot{\timeind-1} &\bm
			 0 \\ \bm 0 & \bm 0 & \bm 0& \ldots & \bm 0&
			 -\transitionmat\timenot{\timeind}
			 &\identitymat_\vertexnum \\ \end{array} \right] \end{align}} \normalsize
			 which is a submatrix of
			 $\extendedobs\timenot{\timeind}$ that results
			 from removing the last block-row, together
			 with the
			 ${Nt+\extendedsamplenum\timenot{\timeind-1}\times
			 Nt+\extendedsamplenum\timenot{\timeind-1}}$
			 matrix
			 $\extendednoobssigma \timenot{\timeind}\define\bdiagnb\{\plantnoisekernelmat\timenot{1}, \observationnoisevar\timenot{1}\identitymat_{\samplenum\timenot{1}}, \plantnoisekernelmat\timenot{2}, \observationnoisevar\timenot{2}\identitymat_{\samplenum\timenot{2}}, \ldots,\plantnoisekernelmat\timenot{t-1},$
			 $ \observationnoisevar\timenot{t-1}\identitymat_{\samplenum\timenot{t-1}}, \plantnoisekernelmat\timenot{t}\}$,
			 which is a submatrix of
			 $\extendedsigma\timenot{\timeind}$ resulting
			 from removing the last block-row and
			 block-column. Now, replace $\timeind$ with
			 $\timeind-1$ and $\timeindp$ with $\timeind$
			 in \eqref{eq:truekfobj} to
			 obtain \begin{align} \extendedgrowingsignalestvec\timegiventimenot{\timeind}{\timeind-1}=\underset{ \extendedgrowingsignalvec\timenot{\timeind}}{\argmin}||\extendedgrowingnoobsvec\timenot{\timeind}
			 - \extendedtrs\timenot{\timeind}\extendedgrowingsignalvec\timenot{\timeind}
			 ||^2_{\extendednoobssigma\timenot{\timeind}} \label{eq:transtruenoobskfobj} \end{align}
			 where
			 $\extendedgrowingnoobsvec\timenot{\timeind} \define[\bm
			 0_\vertexnum\transpose,\observationvec\transpose\timenot{1}, \bm
			 0_\vertexnum\transpose,\observationvec\transpose\timenot{2}, \bm
			 0_\vertexnum\transpose,\ldots,$
			 $ \observationvec\transpose\timenot{\timeind-1},\bm
			 0_\vertexnum\transpose]\transpose \in\rfield^{\vertexnum\timeind+\extendedsamplenum\timenot{\timeind-1}}$ is a submatrix of
			 $\extendedgrowingobsvec\timenot{\timeind}$
			 that results from removing its last
			 block-row.  In this case,
			 $\extendedgrowingsignalestvec\timegiventimenot{\timeind}{\timeind-1}$
			 in~\eqref{eq:transtruenoobskfobj}
			 corresponds to the least-squares solution
			 to \eqref{eq:systemofeq} after removing the
			 last $\samplenum\timenot{\timeind}$ equations,
and
			 can be obtained in 
			  closed form as \begin{align} \extendedgrowingsignalestvec\timegiventimenot{\timeind}{\timeind-1}=
			 (\extendedtrs\transpose \timenot{\timeind}\extendednoobssigma\inv\timenot{\timeind}\extendedtrs \timenot{\timeind})\inv\extendedtrs\transpose \timenot{\timeind}\extendednoobssigma\inv\timenot{\timeind} \extendedgrowingnoobsvec \timenot{\timeind}.  \label{eq:closedformtotminone} \end{align}
}

\change{The rest of the proof utilizes \eqref{eq:closedformtot} and
\eqref{eq:closedformtotminone} to express
$\signalestvec\timegiventimenot{\timeind}{\timeind} $ in terms of
$\signalestvec\timegiventimenot{\timeind}{\timeind-1} $, and
$\signalestvec\timegiventimenot{\timeind}{\timeind-1} $ in terms of
$\signalestvec\timegiventimenot{\timeind-1}{\timeind-1} $. To this
end, define 
$\selectormatrix\timenot{\timeind}\define \canonicalvec{\timeind}{\timeind}\transpose\otimes \identitymat_{\vertexnum}
$, which can be used to select the last $\vertexnum\times\vertexnum$
block-row or block-column of a matrix,
 as well as		
}
			\begin{myitemize} \myitem\cmt{error covariance
				updates} \change{\begin{align} \errormat\timegiventimenot{t}{t-1} \define\selectormatrix\timenot{\timeind}
				(\extendedtrs\transpose \timenot{\timeind}\extendednoobssigma\inv\timenot
				{\timeind}\extendedtrs \timenot{\timeind})\inv\selectormatrix\transpose \timenot{\timeind} \label{eq:errorpred} \end{align}and \begin{align} \errormat\timegiventimenot{t}{t}\define\selectormatrix\timenot{\timeind}
				(\extendedobs\transpose \timenot{\timeind}\extendedsigma\inv\timenot{\timeind}\extendedobs \timenot{\timeind})\inv\selectormatrix\transpose\timenot{\timeind}, \label{eq:errorcor} \end{align}
				which respectively correspond to the
				bottom right
				$\vertexnum\times\vertexnum$ blocks of
				$\extendedfactornoobs\timenot{\timeind} \define\extendedtrs\transpose \timenot{\timeind}\extendednoobssigma\inv\timenot
				{\timeind}\extendedtrs \timenot{\timeind}$
				and
				$\extendedfactor\timenot{\timeind}\define \extendedobs\transpose \timenot{\timeind}\extendedsigma\inv\timenot{\timeind} \extendedobs \timenot{\timeind}
				$.  Expressions \eqref{eq:errorcor}
				and \eqref{eq:errorpred} will be used
				next to express
				$ \errormat\timegiventimenot{t}{t-1} $
				in terms of
				$ \errormat\timegiventimenot{t-1}{t-1}
				$, and
				$ \errormat\timegiventimenot{t}{t} $
				in terms of
				$ \errormat\timegiventimenot{t}{t-1}
				$. 
}

\change{Assume for simplicity that $\extendedsigma\timenot{\timeind}$ and
				$\extendednoobssigma\timenot{\timeind}$
				equal the identity matrices of
				appropriate sizes, although the proof
				easily carries over to arbitrary
				positive definite matrices
				$\extendedsigma\timenot{\timeind}$ and
				$\extendednoobssigma\timenot{\timeind}$. Note
				that \begin{align} \extendedfactornoobs\timenot{\timeind}= \left[ \begin{array}{cc} \extendedfactor\timenot{\timeind-1}+ \verticalincrement\transpose\timenot{\timeind} \verticalincrement\timenot{\timeind}
				& \verticalincrement\transpose\timenot{\timeind}\\ \verticalincrement\timenot{\timeind}
				& \identitymat_\vertexnum \\ \end{array} \right] \label{eq:blockmat} \end{align}
				where
				$\verticalincrement\timenot{\timeind}\define
				-\transitionmat\timenot{\timeind}\selectormatrix \timenot{\timeind-1}\in \rfield^{\vertexnum\times\vertexnum(\timeind-1)}$
				and observe that
				$\errormat\timegiventimenot{t}{t-1}$
				is the bottom right
				$\vertexnum\times\vertexnum$ block of
				$ \extendedfactornoobs\inv\timenot{\timeind}$.
				Thus, applying block matrix inversion
				to \eqref{eq:blockmat}
				yields \begin{align}&\errormat\timegiventimenot{t}{t-1}= \selectormatrix\timenot{\timeind} \extendedfactornoobs\inv\timenot{\timeind} \selectormatrix\transpose \timenot{\timeind}\nonumber\\
				&={\big(}\identitymat_\vertexnum
				-\verticalincrement\timenot{\timeind} \big(\extendedfactor\timenot{\timeind-1}+ \verticalincrement\transpose\timenot{\timeind} \verticalincrement\timenot{\timeind})\inv \verticalincrement\transpose\timenot{\timeind}
				{\big)}\inv.  \label{eq:helpresultforerrormat} \end{align}
				Moreover,  the matrix inversion
				lemma yields, \begin{align}
				&(\extendedfactor\timenot{\timeind-1}+ \verticalincrement\transpose\timenot{\timeind} \identitymat_\vertexnum \verticalincrement\timenot{\timeind}\big)\inv= \extendedfactor\inv\timenot{\timeind-1}
				-\extendedfactor\inv\timenot{\timeind-1}\times \nonumber\\
				& \verticalincrement\transpose\timenot{\timeind} \big(\identitymat_\vertexnum+ \verticalincrement\timenot{\timeind} \extendedfactor\inv\timenot{\timeind-1} \verticalincrement\transpose\timenot{\timeind}\big)\inv \verticalincrement\timenot{\timeind} \extendedfactor\inv\timenot{\timeind-1} \label{eq:shurresult}.  \end{align}
				Substituting \eqref{eq:shurresult}
				into \eqref{eq:helpresultforerrormat},
				applying the definition of
				$\verticalincrement\timenot{\timeind}$,
				and using~\eqref{eq:errorcor} to
				identify
				$\errormat\timegiventimenot{t-1}{t-1}$
				enables one to express
				$ \errormat\timegiventimenot{t}{t-1}$
			 	in terms of
				$\errormat\timegiventimenot{t-1}{t-1}$ as
\begin{align} \errormat\timegiventimenot{t}{t-1}= \identitymat_\vertexnum+\transitionmat\timenot{\timeind} \errormat\timegiventimenot{t-1}{t-1} \transitionmat\transpose\timenot{\timeind}.  \label{eq:errormatnoobs} \end{align}
}

\change{
On the other hand, to express $ \errormat\timegiventimenot{t}{t}$ in
			 	terms of
			 	$\errormat\timegiventimenot{t}{t-1}$,
			 	note that
			 	$\extendedobs\timenot{\timeind}=[ \extendedtrs\transpose\timenot{\timeind}, \horizontalincrement\transpose \timenot{\timeind}]\transpose$,
			 	where
			 	$\horizontalincrement\timenot{\timeind}\define \samplemat\timenot{\timeind}\selectormatrix\timenot{\timeind} \in \rfield^{\samplenum\timenot{\timeind}\times\vertexnum\timeind}$.
			 	Therefore, \begin{align} \extendedfactor\timenot{\timeind}=& \nonumber
				 \extendedobs\transpose \timenot{\timeind} \extendedobs \timenot{\timeind}\\=&
\extendedtrs\transpose \timenot{\timeind}\extendedtrs \timenot{\timeind}+ \horizontalincrement\transpose \timenot{\timeind}\horizontalincrement\timenot{\timeind}\nonumber
\\=&
 \extendedfactornoobs\timenot{\timeind}+ \horizontalincrement\transpose \timenot{\timeind}\horizontalincrement\timenot{\timeind}.  \label{eq:extendedclosedform} \end{align}
			 	Applying the matrix inversion lemma to~\eqref{eq:extendedclosedform} 
			 	yields \begin{align}
				\label{eq:helpshur}
			 	\extendedfactor\inv\timenot{\timeind}=& \extendedfactornoobs\inv\timenot{\timeind}- \extendedfactornoobs\inv\timenot{\timeind}\horizontalincrement \transpose\timenot{\timeind}\times\\
			 	& \big(\identitymat_{\samplenum \timenot{\timeind}}+ \horizontalincrement \timenot{\timeind} \extendedfactornoobs\inv\timenot{\timeind} \horizontalincrement\transpose\timenot{\timeind}\big)\inv
\nonumber
				\horizontalincrement\timenot{\timeind}
				\extendedfactornoobs\inv\timenot{\timeind}.
				\end{align}
Substituting the definition of 
				$\horizontalincrement\timenot{\timeind}$
				into  \eqref{eq:helpshur} leads to
				\begin{align}
				&\selectormatrix\timenot{\timeind} 
				\extendedfactor\inv\timenot{\timeind}= 
				\selectormatrix\timenot{\timeind} 
				\extendedfactornoobs\inv\timenot{\timeind}-
				\selectormatrix\timenot{\timeind} 
				\extendedfactornoobs\inv\timenot{\timeind}\selectormatrix 
				\transpose\timenot{\timeind}\samplemat\transpose\timenot 
				{\timeind}\times\nonumber\\
				&
				\big(\identitymat_{\samplenum 
					\timenot{\timeind}}+ \samplemat\timenot 
				{\timeind}\selectormatrix 
				\timenot{\timeind} 
				\extendedfactornoobs\inv\timenot{\timeind} 
				\selectormatrix\transpose\timenot{\timeind}
				\samplemat\transpose\timenot 
				{\timeind}\big)\inv 
				\samplemat\timenot 
				{\timeind}\selectormatrix\timenot{\timeind}
				\extendedfactornoobs\inv\timenot{\timeind}\nonumber\\&= 
				\selectormatrix\timenot{\timeind} 
				\extendedfactornoobs\inv\timenot{\timeind}-
				\errormat\timegiventimenot{t}{t-1}\samplemat\transpose\timenot 
				{\timeind}\times\nonumber\\
				&~~~
				\big(\identitymat_{\samplenum 
					\timenot{\timeind}}+ \samplemat\timenot 
				{\timeind}
				\errormat\timegiventimenot{t}{t-1}
				\samplemat\transpose\timenot 
				{\timeind}\big)\inv 
				\samplemat\timenot 
				{\timeind}\selectormatrix\timenot{\timeind}
				\extendedfactornoobs\inv\timenot{\timeind}\nonumber\\
				&=(\identitymat_\vertexnum-
				\kalmangainmat\timenot{t}\samplemat\timenot{t})
				\selectormatrix\timenot{\timeind}
				\extendedfactornoobs\inv\timenot{\timeind}
				\label{eq:helpshurimmediate}
				\end{align}
where the second equality follows from \eqref{eq:errorpred}, and the
				third from 
\begin{align}
\label{eq:simplifiedkalmangainmat}
				\kalmangainmat\timenot{t}
				\define\errormat\timegiventimenot{t}{t-1}\samplemat\transpose\timenot{t}
				(\identitymat_{\samplenum 
					\timenot{\timeind}}+\samplemat\timenot{t} 
				\errormat\timegiventimenot{t}{t-1}
				\samplemat\transpose\timenot{t})^{-1}.
\end{align}
				Finally, multiplying both sides
				of~\eqref{eq:helpshurimmediate} with
				$\selectormatrix\transpose\timenot{\timeind}
				$ and using~\eqref{eq:errorcor} to
				identify
				$\errormat\timegiventimenot{t}{t}$
				enables one to express
				$ \errormat\timegiventimenot{t}{t}$ in
				terms of
				$ \errormat\timegiventimenot{t}{t-1}$
				as \begin{align}\label{eq:updateerrormattt} \errormat\timegiventimenot{t}{t}=(\identitymat_\vertexnum
				-\kalmangainmat\timenot{t}\samplemat\timenot{t}) \errormat\timegiventimenot{t}{t-1}.  \end{align}
				If $\extendedsigma\timenot{\timeind}$
				and
				$\extendednoobssigma\timenot{\timeind}$
				are not identity matrices, then one
				obtains
				\begin{align}
					\errormat\timegiventimenot{t}{t-1}= 
					\plantnoisekernelmat\timenot{t}+\transitionmat\timenot{\timeind} 
					\errormat\timegiventimenot{t-1}{t-1} \transitionmat\transpose\timenot{\timeind}
					\label{eq:trueerrormat}
				\end{align}
				instead of \eqref{eq:errormatnoobs},
				and 
				\begin{align}
					\kalmangainmat\timenot{t}
					=\errormat\timegiventimenot{t}{t-1} \samplemat\transpose\timenot{t}
					(\observationnoisevar\timenot{\timeind}\identitymat_{\samplenum 
						\timenot{\timeind}}
					+\samplemat\timenot{t}\errormat 
					\timegiventimenot{t}{t-1}\samplemat\transpose\timenot{t})^{-1}
					\label{eq:truekalmangain}
				\end{align} 
				instead
				of \eqref{eq:simplifiedkalmangainmat},
				whereas \eqref{eq:updateerrormattt}
				remains the same. These equations
				are precisely those in 
				steps \ref{step:predictionerror}, \ref{step:gain}
				and \ref{step:correctionerror} of
				Algorithm~\ref{algo:kalmanfilter}.  
}

\myitem\cmt{state
				updates}\change{To obtain the rest of the
 steps,  set $\timeind$ to
 $\timeind-1$ and $\tau$ to $\timeind$ in
 \eqref{eq:predictiontau} to obtain
				\begin{align}
				\signalestvec\timegiventimenot{\timeind}{\timeind-1}= 
				\transitionmat\timenot{\timeind}
				\signalestvec\timegiventimenot{\timeind-1}{\timeind-1}	
				\label{eq:transitionproof}		
				\end{align}
			    which coincides with step~\ref{step:prediction} of 
			    Algorithm~\ref{algo:kalmanfilter}.
 Finally, since $\signalestvec\timegiventimenot{\timeind}{\timeind}$
			    is the last block vector of $\extendedgrowingsignalestvec 
			    \timegiventimenot{\timeind}{\timeind}$, then
				\begin{align}
\nonumber
				\signalestvec\timegiventimenot{\timeind}{\timeind}
				\define&\selectormatrix\timenot{\timeind} 
				\extendedgrowingsignalestvec 
				\timegiventimenot{\timeind}{\timeind}\\
				=&\selectormatrix\timenot{\timeind} 
				\extendedfactor\inv\timenot{\timeind}\extendedobs\transpose 
				\timenot{\timeind}\extendedsigma\inv\timenot{\timeind}
				\extendedgrowingobsvec\timenot{\timeind}\nonumber\\ 
				=&(\identitymat-\kalmangainmat\timenot{t}
				\samplemat\timenot{t})\selectormatrix\timenot{\timeind}
				\extendedfactornoobs\inv\timenot{\timeind} 
				\extendedobs\transpose 
				\timenot{\timeind}\extendedsigma\inv\timenot{\timeind}\extendedgrowingobsvec
				\timenot{\timeind}
\label{eq:fhtt}
				\end{align}
				where the second equality follows
			    from \eqref{eq:closedformtot} and the
			    third from~\eqref{eq:helpshurimmediate}.
From the definitions of $\extendedobs	\timenot{\timeind}$,
				$\extendedsigma\timenot{\timeind}$ and $\extendedgrowingobsvec
				\timenot{\timeind}$, one obtains that
\begin{align}
\label{eq:asigmapsi}
\extendedobs\transpose 
				\timenot{\timeind}\extendedsigma\inv\timenot{\timeind}\extendedgrowingobsvec
				\timenot{\timeind}
= \extendedtrs\transpose 
				\timenot{\timeind}\extendednoobssigma\inv\timenot{\timeind} 
				\extendedgrowingnoobsvec
				\timenot{\timeind}+\frac{1} 
				{\observationnoisevar\timenot{\timeind}} 
				\horizontalincrement\transpose\timenot{\timeind}
				\observationvec\timenot{\timeind}.
\end{align}
Substituting \eqref{eq:asigmapsi} into \eqref{eq:fhtt} yields
\begin{align}
\signalestvec\timegiventimenot{\timeind}{\timeind}=&(\identitymat-\kalmangainmat\timenot{t}
				\samplemat\timenot{t})\selectormatrix\timenot{\timeind} \extendedfactornoobs\inv\timenot{\timeind}\times\nonumber\\
				&(\extendedtrs\transpose \timenot{\timeind}\extendednoobssigma\inv\timenot{\timeind} \extendedgrowingnoobsvec \timenot{\timeind}+\frac{1}
				{\observationnoisevar\timenot{\timeind}} \horizontalincrement\transpose\timenot{\timeind} \observationvec\timenot{\timeind}
				)\nonumber\\
				=&(\identitymat-\kalmangainmat\timenot{t} \samplemat\timenot{t})(\signalestvec\timenot{t|t-1}\nonumber\\
				&+\frac{1}
				{\observationnoisevar\timenot{\timeind}} \errormat\timegiventimenot{t}{t-1} \samplemat\transpose\timenot{\timeind} \observationvec\timenot{\timeind}
				)\nonumber\\
				=&\signalestvec\timenot{t|t-1}+ \kalmangainmat\timenot{t}(\observationvec\timenot{t}- \samplemat\timenot{t}\signalestvec\timenot{t|t-1}) \label{eq:correctionproof} \end{align}
				where the second equality follows
				from~\eqref{eq:closedformtotminone},
				$\signalestvec\timegiventimenot{\timeind}{\timeind-1}
				=\selectormatrix\timenot{\timeind} \extendedgrowingsignalestvec \timegiventimenot{\timeind}{\timeind-1}$
				and~\eqref{eq:errorpred}; whereas the
				third follows from 	\begin{align}
			 (\identitymat_{\vertexnum}-	\kalmangainmat\timenot{t}\samplemat
			\timenot{t})\errormat\timegiventimenot{t}{t-1}\samplemat
			\transpose\timenot{t}
				=\observationnoisevar\timenot{\timeind}
					\kalmangainmat\timenot{t}
				\label{eq:altkalmangain}
				\end{align} which results from rearranging the terms in~\eqref{eq:truekalmangain}.
				Noting that expression \eqref{eq:correctionproof}
				coincides with
				step~\ref{step:correction} of
				Algorithm~\ref{algo:kalmanfilter}
				concludes the proof.  }\end{myitemize}
		\end{myitemize}
	\end{myitemize}

\bibliography{my_bibliography}
\bibliographystyle{IEEEbib}

\end{document}